\documentclass[runningheads]{llncs}
\usepackage{graphicx}
\usepackage{amsmath,amssymb} % define this before the line numbering.
\usepackage{color}
\usepackage{times}
\usepackage{amsmath}
\usepackage{amssymb}
\usepackage{url}
\usepackage{graphicx}
\usepackage{epstopdf}
\usepackage{algorithm2e}
\usepackage{algorithmicx}
\usepackage[width=122mm,left=12mm,paperwidth=146mm,height=193mm,top=12mm,paperheight=217mm]{geometry}

\usepackage{colortbl}

\newcommand{\red}[1]{#1}
\newcommand{\blue}[1]{#1}

\newcommand{\mbr}[1]{\mathrm{\mathbf{#1}}}
\newcommand{\etal}{\emph{et al.~}}

\newcommand{\bb}[1]{\boldsymbol{\mathrm{#1}}}

\newcommand{\zz}{\mbr{z}}

\usepackage [english]{babel}
\usepackage [autostyle, english = american]{csquotes}
\MakeOuterQuote{"}
\graphicspath{{./}}

\begin{document}
	%\linenumbers
	\pagestyle{headings}
	\mainmatter
	\def\ECCV16SubNumber{463}  % Insert your submission number here
	
	\title{FPGA system for real-time computational extended depth of field imaging using phase aperture coding} % Replace with your title
	
	\titlerunning{ECCV-16 submission ID \ECCV16SubNumber}
	
	\authorrunning{ECCV-16 submission ID \ECCV16SubNumber}
	
	    \author{Tal Remez
	    	\and
	    	Or Litany
	    	\and
	    	Shachar Yoseff
	    	\and
	    	Harel Haim
	    	\and
	    	Alex Bronstein
	    }	
	\institute{Tel-Aviv University\\
		Electrical Engineering Department, Tel Aviv University, Tel Aviv, Israel\\}
	
	\maketitle
	
	\begin{abstract}
		We present a proof-of-concept end-to-end system for computational extended depth of field (EDOF) imaging. The acquisition is performed through a phase-coded aperture implemented by placing a thin wavelength-dependent optical mask inside the pupil of a conventional camera lens, as a result of which, each color channel is focused at a different depth. The reconstruction process receives the raw Bayer image as the input, and performs blind estimation of the output color image in focus at an extended range of depths using a patch-wise sparse prior. We present a fast non-iterative reconstruction algorithm operating with constant latency in fixed-point arithmetics and achieving real-time performance in a prototype FPGA implementation. The output of the system, on simulated and real-life scenes, is qualitatively and quantitatively better than the result of clear-aperture imaging followed by state-of-the-art blind deblurring.
		
		\keywords{\red{EDOF, sparse coding, FPGA, real-time.}}
	\end{abstract}
	
	%%%%%%%%% BODY TEXT
	\section{Introduction}
	The increasing popularity of phone cameras combined with the affordability of high-quality CMOS sensors have transformed digital photography into an integral part of the way we communicate. While being the principal technology driver for miniature cameras, the smartphone market is also one of the most challenging arenas for digital imaging. The quality of a digital image is determined by the quality of the optical system and the image sensor. With the increase in pixel number and density, imaging system resolution is now mostly bound by optical system limitations. Since form factor challenges imposed by smart phone designs make it very difficult to improve the image quality via standard optical solutions, many R\&D activities in the field in recent years have shifted to the domain of computational imaging.
	The need to acquire high-quality images and videos of moving low-light scenes through a miniature lens and the fact that traditional autofocusing mechanisms are costly, slow, and unreliable in low-light conditions render acute the tradeoff between the aperture size (F\#) and the depth of field (DOF) of the optical system. 
	
	Imaging with limited DOF brings forth the challenge of restoration of out-of-focus (OOF) images -- a  notoriously ill-posed problem due to information loss in the process. There exists a wealth of literature dedicated to purely computational approaches to image deblurring and deconvolution \cite{Almeida2010,krishnan2011blind,Shan2008,Starck2002}. The dominant models, increasingly popular in recent years, are flavors of sparse and redundant representations \cite{Couzinie-Devy2011,Hu2010}, which have been proved as a powerful tool for image processing, compression, and analysis. It is now well-established that small patches from a natural image can be represented as a linear combination of only a few atoms in an appropriately constructed over-complete (redundant) dictionary. This constitutes a powerful prior that has been successfully employed to regularize numerous otherwise ill-posed image processing and restoration tasks \cite{elad2006image,Fadili2009}.
	
	Existing image deblurring techniques assume that the blurring kernel is known \cite{Couzinie-Devy2011} or use an expectation maximization-type of approach to blindly estimate the kernel from the data, either parametrically when the point spread function  (PSF) of the system is well-characterized and only the scene depth is unknown, or non-parametrically making less assumptions about the optics \cite{Almeida2010,Hu2010,krishnan2011blind,Shan2008}. Using conventional optics, information about sharp image features is irreversibly lost, posing inherent limitations on any computational technique.
	
	\blue{
		These limitations can be overcome by manipulating the image acquisition process. Some well-known methods produced a constant PSF kernel for all depths of field using a wave-front coding phase mask \cite{dowski1995extended,cossairt2010diffusion}. Similar methods for depth-invariant PSF utilize focal sweep \cite{nagahara2008flexible} or use uncorrected lens as a type of spectral focal sweep \cite{cossairt2010spectral}. Having a single known PSF, the image can be recovered using non-blind deconvolution methods. These techniques reduce image contrast and are very sensitive to noise and therefore not suitable for high quality image applications.
		
		Other related techniques use an amplitude coded mask \cite{Levin2007} or a color-dependent ring mask \cite{chakrabarti2012depth} such that objects at different depth will exhibit a distinctive spatial structure which improves the image restoration process. The main drawback of those methods is the fact that the light efficiency is only about 50\% in \cite{Levin2007} and 60\% in \cite{chakrabarti2012depth}, making them unsuitable for low light conditions.
	}
	
	A recent trend in the literature \cite{haim2014multi} proposes %novel approach initiated in our group proposed
	the use of wavelength-dependent phase aperture coding, whereby one gets different responses for the red, green and blue channels, resulting in the simultaneous acquisition of three perfectly registered images, each with a different out-of-focus characteristic. Under the assumption that sharp edges and other high-frequency features characterizing a well-focused image are present in all color channels, the joint analysis of the three channels acquired through a phase mask located in pupil, allows to extend the system DOF without significant prior knowledge about the blurring kernel. \blue{This type of mask can be easily assembled in most conventional systems with minimal modification and provides a light efficiency of up to 99\% using a coated glass plate. }

	\textbf{Contributions}
	\red{ presented in this work are threefold: 
		\begin{enumerate}
			\item We present a learning-based algorithm for computational EDOF using phase aperture coding. 	
			\item We demonstrate empirically the superiority of our algorithm over previous methods in reconstruction quality at a significantly lower runtime. 
			\item We introduce a full end-to-end hardware system including a FPGA implementation of our algorithm.
		\end{enumerate}
	}
	
	The rest of the paper is organized as follows:   Section 2 details the image formation model in phase-coded aperture acquisition. Section 3 formulates the reconstruction optimization problem, and Section 4 details the fast algorithm for its approximate solution. Section 5 describes the FPGA system hosting the reconstruction process. Experimental evaluation of the system on synthetic and real scenes, and its comparison to clear-aperture imaging following by standard deconvolution is presented in Section 6. A brief discussion in Section 7 concludes the paper. 
	
	%-------------------------------------------------------------------------
	\section{Phase-Coded Aperture Imaging}
	An imaging system acquiring an out-of-focus object suffers from aberrations, in particular blur, that degrade the image quality resulting in low contrast, loss of sharpness and even loss of information. In digital systems, image blur (main OOF effect) is not observable as long as the image size of a point source in the object plane is smaller than the pixel size in the detector plane. 
	The OOF error is analytically a wave-front error, expressed as a quadratic phase error in the pupil plane \cite{Goodman1996}. In case of a circular aperture with radius \textit{R}, we define the defocus parameter as

	\begin{equation}
		\begin{split}
			\psi
			&= \frac{{\pi {R^2}}}{\lambda }\left( {\frac{1}{{{z_\mathrm{o}}}} + 
				\frac{1}{{{z_{\mathrm{img}}}}} - \frac{1}{f}} \right)
			= \frac{{\pi {R^2}}}{\lambda }\left( {\frac{1}{{{z_\mathrm{img}}}} - \frac{1}{{{z_\mathrm{i}}}}} \right)\\
			&= \frac{{\pi {R^2}}}{\lambda }\left( {\frac{1}{{{z_\mathrm{o}}}} - \frac{1}{{{z_\mathrm{n}}}}} \right),
		\end{split}
	\end{equation}
	where  $z_\mathrm{img}$ is the sensor plane location of an object in the nominal position  $z_\mathrm{n}$,  $z_\mathrm{i}$ is the ideal image plane for an object located at  $z_\mathrm{o}$ , and  $\lambda$  is the optical wavelength.
	The defocus parameter  $\psi$ measures the maximum phase error at the aperture edge.  For $\psi>1$, the image will experience contrast loss, while for $\psi>4$ it will experience information loss and even reversal of contrast at some frequencies, as can be observed in Figure \ref{MTF}(a).
	For a circular clear aperture, the cut-off spatial frequency due to diffraction limit is given by  $f_{\mathrm{c}}=\frac{2R}{\lambda z_{\mathrm{i}}}$. Increasing the aperture size R is hence a double-edged sword: it results in higher $f_{\mathrm{c}}$ and hence improved contrast and higher signal-to-noise ratio (SNR) due to the increase in the amount of collected light; however, this comes at the cost of a steeper increase in $\psi$ as a function of the deviation of the object depth $z_\mathrm{o}$ from the nominal depth $z_\mathrm{n}$, hence, lower DOF. comes at the cost of diminishing the DOF. 
	
	\red{
		\begin{figure}[tb]
			\centering
			\begin{tabular}{ c c}
				\includegraphics[width = 0.45\columnwidth]{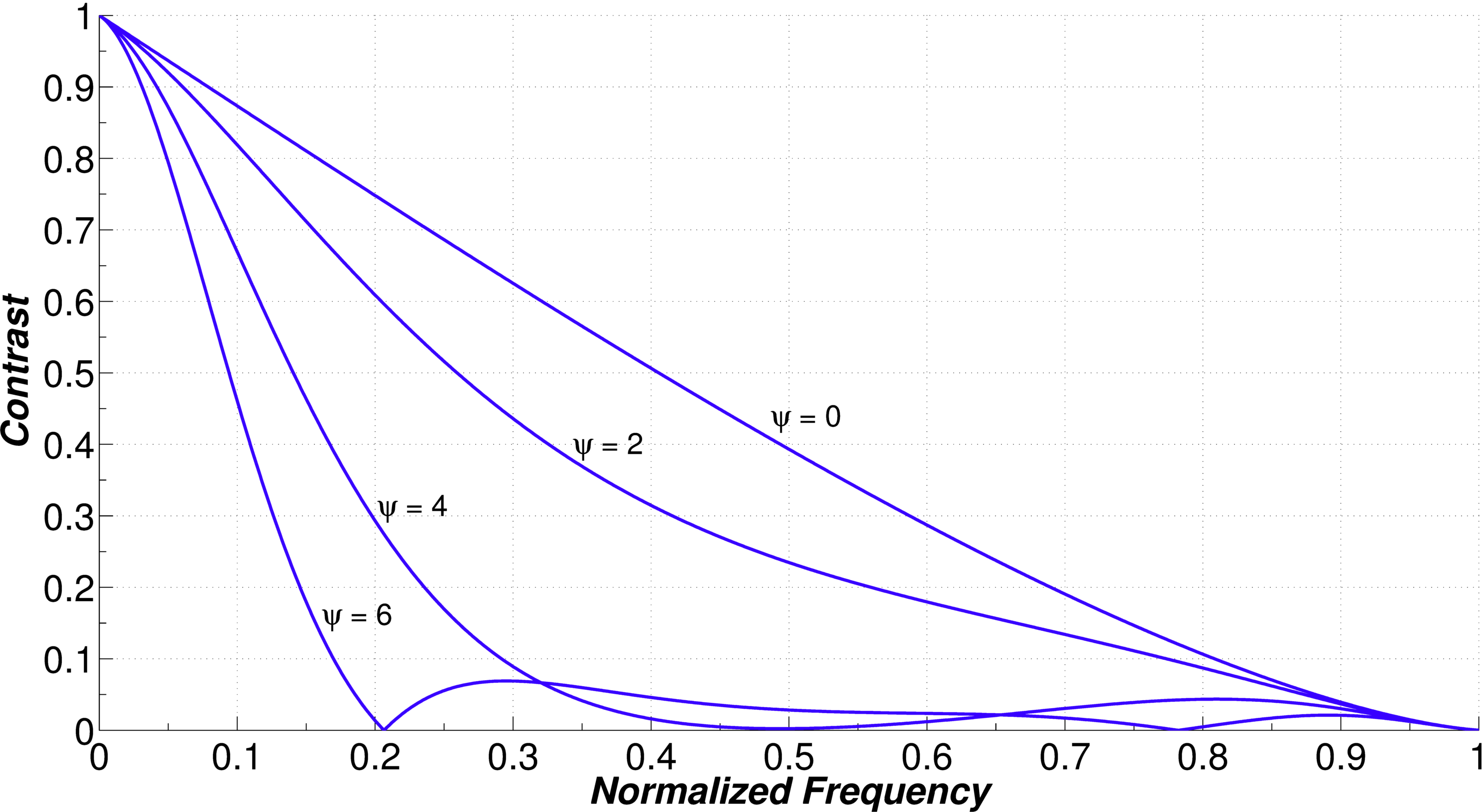} &
				\includegraphics[width = 0.45\columnwidth]{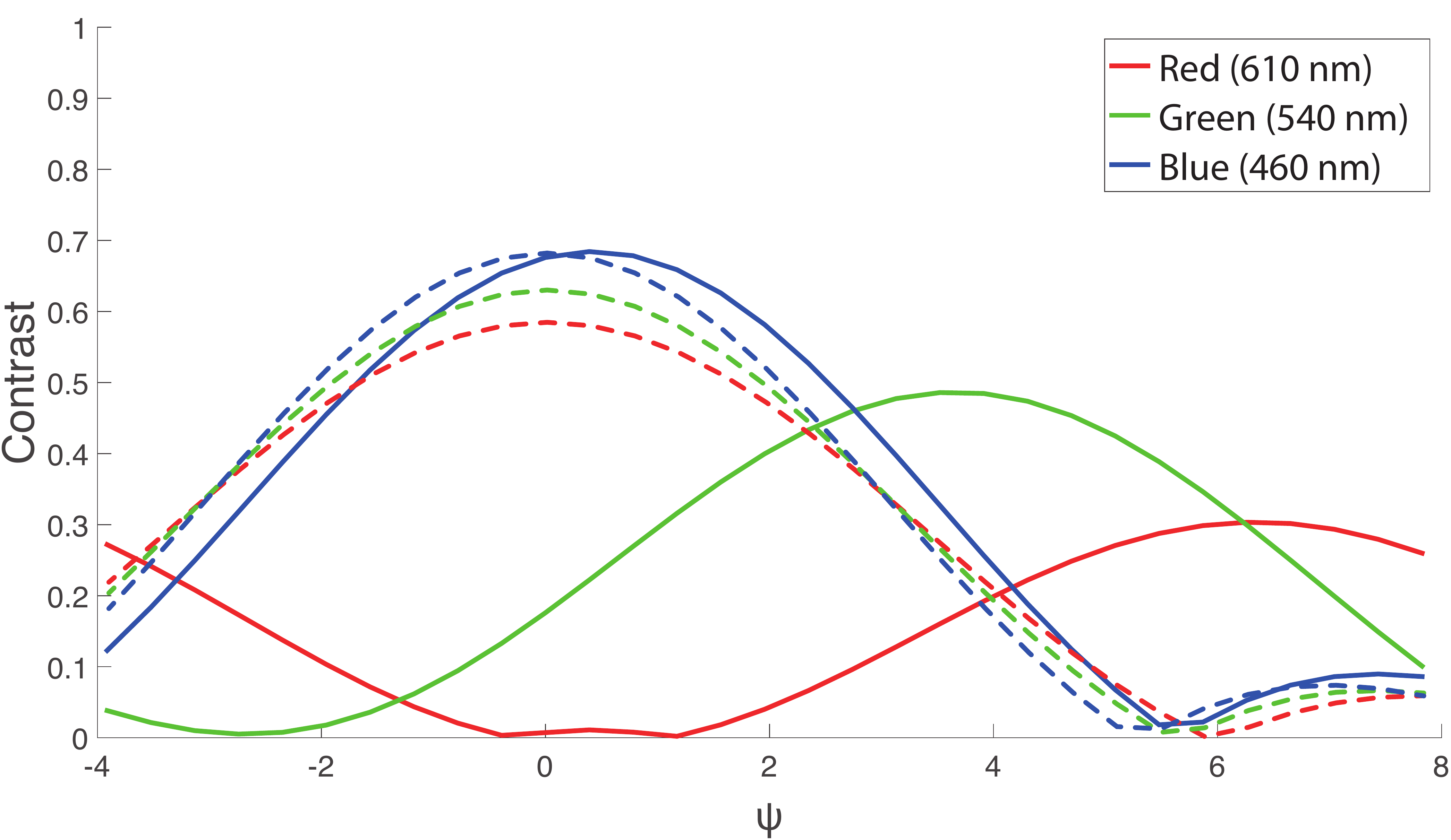} \\	
				(a) & (b) \\
			\end{tabular}
			\caption{\small \textbf{Spatial frequency response and color channel separation} (a) Optical system response to spatial frequency for different $\psi$ values. (b) Comparison between contrast levels for a single spatial frequency $f_{\mathrm{c}}/4$ as a function of $\psi$ with clear aperture (dotted) and a phase mask (solid).}
			\label{MTF}
		\end{figure}
	}
	
	Radially symmetric binary optical phase masks have been proposed for overcoming limitations set by OOF imaging \cite{milgrom2010pupil}. Milgrom \etal \cite{milgrom2010novel} proposed the use of a special RGB phase mask that exhibits significantly different response in the three major color channels R, G and B. It has been shown that each channel provides best performance for different depth regions, so that the three channels jointly provide an extended DOF. Based on Milgrom's mask, a similar mask was designed in \cite{Haim2015} specifically to further increase the diversity between the color channels such that each depth region will be as unique as possible. This effect has been the key element of developing this new type of computational camera. Figure \ref{MTF}(b) shows the diversity between the color channels for different depths ($\psi$  values) when using a clear aperture (dotted plot) and when using the phase mask from \cite{Haim2015} (solid plot).  The contrast levels are generally somewhat lower when using the phase mask, for the "in focus" condition, but it exhibits higher contrast at larger DOF, providing separation of the response for the three color channels, an effect that will later be used for restoring the contrast levels to their nominal values.
	
	%\paragraph{Image formation model.} 
	Assuming approximately constant depth and, hence, the same unknown $\psi$ for all pixels in the vicinity of a spatial location $\bb{\xi}$ in the sensor plane, the image formation model can be described as the following convolution
	\begin{eqnarray*}
		y^{\mathrm{R}}(\bb{\xi})& =& (h^{\mathrm{R}}_\psi \ast x^{\mathrm{R}})(\bb{\xi}) + \eta^{\mathrm{R}}(\bb{\xi})  \nonumber\\
		y^{\mathrm{G}}(\bb{\xi}) &= & ( h^{\mathrm{G}}_\psi \ast x^{\mathrm{G}})(\bb{\xi}) + \eta^{\mathrm{G}}(\bb{\xi})  \nonumber\\
		y^{\mathrm{B}}(\bb{\xi}) &=& ( h^{\mathrm{B}}_\psi \ast x^{\mathrm{B}} )(\bb{\xi}) + \eta^{\mathrm{B}} (\bb{\xi}),
	\end{eqnarray*}
	where $x^\ast,*=R,G,B,	$ denote the three color channels of the ideal (latent) in-focus image, $y^\ast$ are the corresponding channels of the OOF image formed in the sensor plane, $h^\ast_\psi$ are the three corresponding depth-dependent blur kernels, and $\eta^\ast$ denote additive sensor noise that can be fairly accurately modeled as white Gaussian at least at reasonable illumination intensities. In vector notation combining the three channels, the expression simplifies to
	\begin{eqnarray}
		\bb{y}(\bb{\xi}) = (\bb{h}_\psi \ast \bb{x})(\bb{\xi}) + \bb{\eta}(\bb{\xi}),
	\end{eqnarray}
	where $\bb{x}$ and $\bb{y}$ denote the three-channel latent in-focus and the observed OOF images, respectively. 
	
	In the majority of conventional image sensors, the acquisition of color information is done through a color filter array (CFA), also known as Bayer filter mosaic, where every pixel captures only one of the red, green or blue channels. A demosaicing process then restores the raw image to a full-resolution color image.
	With some abuse of notation, CFA can be introduced to our image formation model as
	\begin{eqnarray}
		y(\bb{\xi}) =\bb{B}(\bb{\xi}) (\bb{h}_\psi \ast \bb{x})(\bb{\xi}) + \bb{\eta}'(\bb{\xi}),
		\label{eq:forward-model}
	\end{eqnarray}
	where $y$ denotes the raw Bayer image, and $\bb{B}(\bb{\xi})$ the location-dependent response of the array filters, which up to cross-talk and vignetting artifacts looks like $\bb{B} = (1,0,0), (0,1,0)$, and $(0,0,1)$ for the R, G, and B color channels, respectively.
	
	\section{EDOF Image Recovery Using a Sparse Prior}
	\label{recovery}
	Sparse representation was proved to be a strong prior for both non-blind \cite{Couzinie-Devy2011} and blind \cite{Almeida2010,Hu2010,krishnan2011blind,Shan2008} image deblurring. A signal $\bb{x} \in \mathbb{R}^{n}$ is said to admit a sparse representation (or, more accurately, approximation) in an $n \times k$ overcomplete $(k>n)$ dictionary $\bb{D}$  if one can find a vector $\bb{z} \in \mathbb{R}^{k}$  with only a few non-zero coefficients, such that $\bb{x} \approx \bb{D} \bb{z}$ e.g. in the $\ell_2$-sense. 
	The sparse representation pursuit problem can be cast as the minimization problem,
	\begin{equation} 	
		\label{sparse_app_eq}
		\hat {\bb{z}} = \mathrm{\arg}\min_{\bb{z}} \|\bb{x}-\bb{D}\bb{z}\|_2^2 + \mu g(\bb{z}),
	\end{equation}
	where $g(\bb{z})$ is a sparsity-inducing prior whose relative importance is governed by the parameter $\mu$.
	
	While priors such as the $\ell_0$ pseudonorm counting  the number of non-zeros in the vector $\bb{z}$
	give rise to  computationally intractable optimization problems, there exist efficient greedy approximation techniques including the orthogonal matching pursuit (OMP) \cite{Chen1998,Tropp2007} or convex relaxation techniques replacing the $\ell_0$ pseudonorm with the $\ell_1$ norm minimization \cite{Elad:2010:SRR:1895005}. 
	The dictionary $\mbr{D}$ can be constructed axiomatically based on image transforms such as DCT or wavelet, or trained (e.g., via $k$-SVD \cite{aharon2006svd}) using a set of representative images. Here, we adopt the latter approach to construct a dictionary representing  $8\times8$ patches from the image, represented as $n=192$-dimensional vectors ($64$ dimensions for each color channel). 
	
	Assuming (unrealistically) a known $\psi$, the forward model (\ref{eq:forward-model}) described in the previous section can be directly incorporated into the data fitting term, resulting in the following optimization problem:
	\begin{equation} 	
		\hat {\bb{z}} = \mathrm{\arg}\min_{\bb{z}} \|\bb{y}- \bb{B} \bb{H}_\psi \bb{D} \bb{z}\|_2^2 + \mu g(\bb{z}),
	\end{equation}
	where $\bb{y}$ is a $64$-dimensional vector representing the patch in the input Bayer image, $\bb{D}$ is the $192 \times k$-dimensional dictionary, and $\bb{z}$ is the $k$-dimensional vector of representation coefficients. The $192 \times 192$ matrix $\bb{H}_\psi$ represents the action of the out-of-focus blur kernel corresponding to $\psi$ to each of the color channels, while $64 \times 192$ matrix $\bb{B}$ describes the action of the CFA on the patch; in conventional CFA designs it is constant as long as the patches are selected from essentially same locations.  
	The reconstructed patch is obtained as $\hat{\bb{x}} = \bb{D} \hat {\bb{z}}$. Defining the blurred dictionary $\bb{D}_\psi = \bb{H}_\psi \bb{D}$, the latter optimization problem can be rewritten as
	\begin{equation} 	
		\hat {\bb{z}} = \mathrm{\arg}\min_{\bb{z}} \|\bb{y}- \bb{B} \bb{D}_\psi \bb{z}\|_2^2 + \mu g(\bb{z}).
	\end{equation}
	Note that the pursuit is performed with respect to the blurred dictionary $\bb{D}_\psi$, while the patch reconstruction from the computed coefficients is performed with the clear dictionary $\bb{D}$. 
	
	In practical applications, the depth and, hence, the out-of-focus parameter $\psi$ in each patch are unknown \emph{a priori}, leading to the blind setting of the above reconstruction problem.
	Many studies dealt with this setting with limited success. For instance, using different iterative processes, one tries to estimate the blurring kernel so that image restoration can be thereafter achieved. Reconstruction processes usually require high computational complexity, which limits their use for many real-time applications. Even then, the assumption of a single blur kernel for the entire image is often unrealistic.
	
	In our setup, the phase coding of the aperture contains hints about $\psi$ through the relation between the color channels.
	To address the blind setting of the problem, we construct a concatenated blurred representation dictionary containing several fixed values of $\psi$,
	$$
	\bb{D}_\Psi = (\bb{D}_{\psi_1}, \dots, \bb{D}_{\psi_q} ),
	$$
	and solving 
	\begin{equation} \label{pursuit_problem}	
		\hat {\bb{z}} = \mathrm{\arg}\min_{\bb{z}} \|\bb{y}- \bb{B} \bb{D}_\Psi \bb{z}\|_2^2 + \mu g(\bb{z}),
	\end{equation}
	for $\bb{z} \in \mathbb{R}^{qk}$.  The reconstruction is performed as before with the clear dictionary concatenated $q$ times, $\hat{\bb{x}} = (\bb{D},\dots,\bb{D}) \hat {\bb{z}}$. In all our experiments, we used $q=8$ and set $\psi_1=1,\dots,\psi_8 = 8$. While an additional prior asserting that the coefficients are localized in one of the sub-dictionaries can be imposed using standard structured sparsity-promoting priors such as the mixed $\ell_{2,1}$ norm, we defer this obvious extension to future study.
	
	The pursuit process chooses elements from the dictionary that best match the input patch, based on the $\ell_2$ distance. Using the RGB phase mask, the blurred dictionary varies strongly for different kernels since the response of the imaging system is very different for each color. An input patch from an imaging system phase-coded aperture will also exhibit different response for each color. The response will be unique for each kernel and therefore the input vector will most likely associate with vectors from the dictionary $\bb{D}_{\psi_i}$ experiencing the same blurring process. 
	
	Comparing our algorithm with a state-of-art algorithm with available public domain code, provided by Krishnan et al. \cite{krishnan2011blind}, \red{ and our implementation of \cite{haim2014multi}}, our process produced superior results when applied to natural images out of the Kodak dataset \cite{franzen2004kodak}. We also run the process on texture images (Colored Brodatz Texture database \cite{brodatz1966textures}) and observed similar performance, meaning that the concatenated dictionary is likely to describe well almost any natural scene.

	\section{Fast Image Reconstruction}
	\label{Fast_Image_Reconstruction}
	As was explained in the previous section, the pursuit problem (\ref{pursuit_problem}) can be posed as a convex optimization problem by choosing $g(\zz) = \|\zz\|_1$, and solved using proximal algorithms such as the iterative shrinkage thresholding algorithm (ISTA) or its accelerated version (FISTA) \cite{beck2009fast}. However, these iterative solvers typically require hundreds of iterations to converge, resulting in prohibitive complexity and unpredictable input-dependent latency, which is unacceptable in real-time applications.
	To overcome this limitation, we follow the approach advocated by \cite{gregor-icml-10}, in which a small number, $T$, of ISTA iterations is unrolled into a feed-forward neural network that subsequently undergoes supervised training on typical inputs, as explained in the sequel.
	A pseudo-code of ISTA is given in Algorithm \ref{algo_ista}, where we denote $\bb{Q} = \frac{1}{\rm{\textit{L}}} \bb{B} \bb{D}_\Psi$, $\mbr{S}=\bb{I} - \frac{1}{\rm{\textit{L}}}\bb{Q}^{\rm{T}}\bb{Q}$, and $\sigma_\theta(x) = \max(|x| - \theta,0)\rm{sign}(x)$ is a two-sided shrinkage function with threshold $\theta = \frac{2\mu}{\rm{\textit{L}}}$ applied element-wise. $\rm{\textit{L}}$ denotes a scalar larger than the largest eigenvalue of $\bb{Q}^{\rm{T}}\bb{Q}$.

	\begin{algorithm}[h!]
		\label{algo_ista}
		\KwIn{Data $\bb{x}$, number of iterations $T$, shrinkage threshold $\mbr{\theta}$, matrices $\bb{D}$, $\bb{Q}$, and $\bb{S}$}
		\KwOut{Reconstructed image $\mbr{\hat{x}}$, \\ \hspace{13 mm}auxiliary variables $\{\mbr{z}_t\}_{t=1}^T$,$\{\mbr{b}_t\}_{t=1}^T$ }
		
		initialize $\mbr{z}_1=\bar{0}$ and $\mbr{b}_1 = \bb{Q}^{\rm{T}} \mbr{x}$
		
		\For{t = 1,2,\dots,T}{
			$\mbr{z}_{t+1} = \sigma_\theta \left( \mbr{b}_{t} \right) $ \\ 
			$\mbr{b}_{t+1} = \mbr{b}_{t} + \mbr{S} \left( \mbr{z}_{t+1} - \mbr{z}_t \right)$\\
		}
		$\mbr{\hat{x}}=\mbr{D}\mbr{z}_T$
		\caption{ISTA algorithm.}		
	\end{algorithm}
	As can be easily interpreted from the ISTA algorithm, the network comprises three types of layers: An initialization layer (denoted as I), which finds the representation of the input signal in the dictionary; several ($T-2$) recurrent middle layers (M) performing gradient step followed by shrinkage; and a final layer (F) which translates the resulting dictionary coefficients to the reconstructed signal. All these types of layers can be realized from the single multi-purpose calculator stage shown in Figure \ref{FPGA_pipline_image} (right) that is amenable for hardware implementation. To get the "I" configuration, we set $\bb{b}_{\mathrm{in}} = \bb{0}$, $\mbr{A} = -\mbr{Q}^{\rm{T}}$, $\mbr{c} = \bb{0}$ and $\bb{\theta} = \bb{0}$. The ``M'' configuration layer is fed by the output of the previous calculator, and the matrix $\bb{A}$ is set to $\bb{S}$. The output is further fed to either another ``M'' layer or to the ``F'' layer. Finally, the ``F'' configuration of the calculator is a reduction into multiplication by the matrix $\mbr{D}$. This is achieved by setting $\bb{b}_{\mathrm{in}} = \bb{0}$, and $\mbr{A} = \bb{D}$.
	
	Supervised training of the network is done by initializing the parameters as detailed above, and then adapting them using a stochastic gradient procedure minimizing the reconstruction error $\mathcal{F}$ of the entire network. We use the following empirical loss 

	\begin{equation}
		\label{general_loss}
		\mathcal{F} = \frac{1}{N}\sum_{n=1}^{N}f(\mbr{x}_n^*,\hat{\mbr{x}}_n) ,
	\end{equation}
	which for a large enough training set, $N$, approximates the expected value of $f$ with respect to the distribution of the ground truth signals $\mbr{x}_n^*$. Here, $\hat{\mbr{x}}_n$ denotes the output of the network, and the loss objective $f$ minimized during the training process is the standard sum of squared differences,
	\begin{equation}
		f = \frac{1}{2}\| \mbr{x}_n^* - \hat{\mbr{x}}_n \|_2^2 .
	\end{equation}

	Similarly to \cite{gregor-icml-10}, the output of the network and the derivatives of the loss with respect to the network parameters are calculated using the standard forward and back propagation approach. Practice shows that the training process allows to reduce the number of layers by about two orders of magnitude while achieving a comparable reconstruction quality. A detailed runtime analysis is presented in Section \ref{Results}.
	
	\section{FPGA Image Reconstruction System}
	\label{sec:FPGA}
	To demonstrate that the proposed image reconstruction process is efficient and is amenable to hardware implementation, we built a prototype FPGA system. An FPGA is a programmable chip containing configurable logic blocks and routing resources, therefore offering a fair amount of flexibility previously only possible with software, along with a hardware-like computational speeds and reliability. While being distinct in many aspects from application-specific integrated chips (ASICs), modern FPGAs are the closest approximation of an ASIC one can get without incurring the costs of custom chip manufacturing.

\begin{figure*}[th]
		\centering
		\includegraphics[width = 1\textwidth]{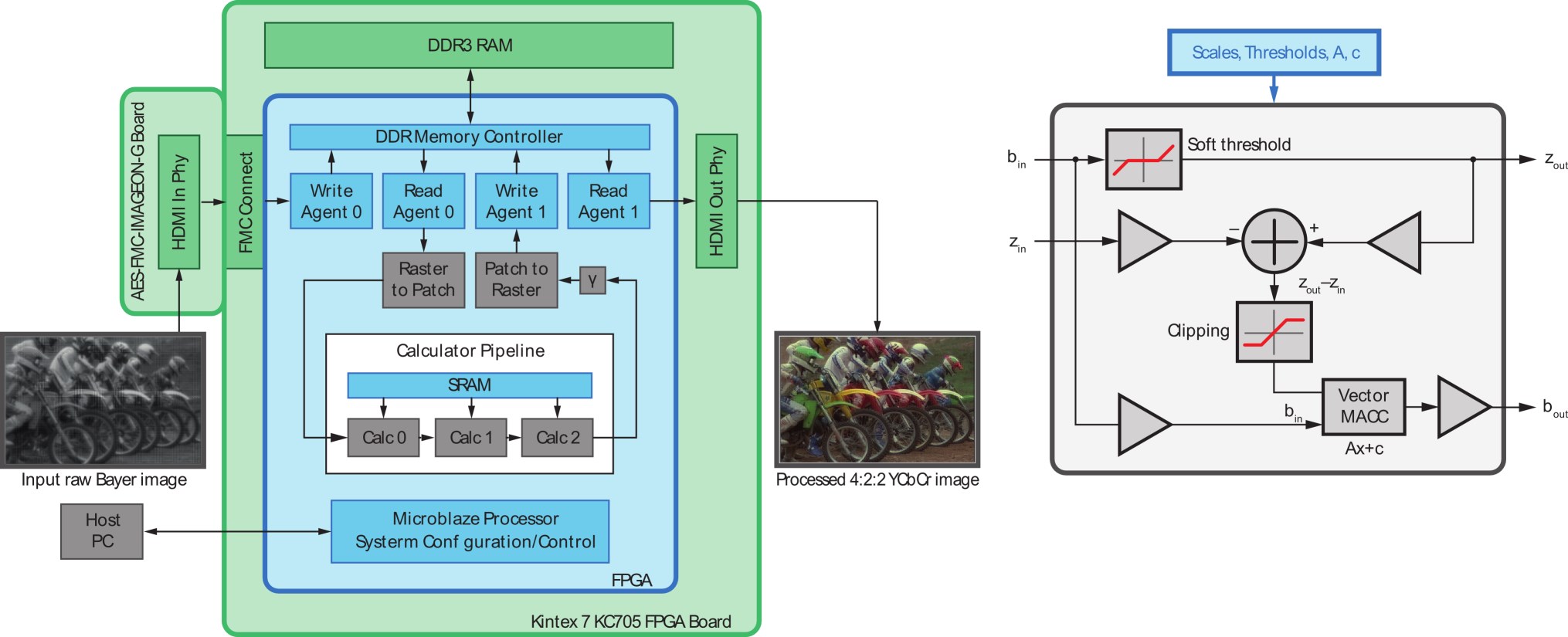}	
		\caption{\small \textbf{Schematic description of the FPGA reconstruction system.} The raw Bayer image from the sensor at 12bit/pixel is passed, through the HDMI input interface daughter board, to the Kintex 7 FPGA chip. The image is buffered in the external DRAM, from where it is fed as a stream of possibly overlapping 8x8 patches to the calculator pipeline comprising of up to eight stages (see detail on the right), implementing the neural network architecture. The output patches in 4:2:2 YCbCr format are average-pooled and buffered in raster order in the DRAM, from where the image is sent over to the HDMI output interface on the FPGA board.  
			The parameters of the calculator stages and other register values controlling the  data flow are stored in the static memory on the chip, into which they are loaded by the host application on system startup.
		}  	
		\label{FPGA_pipline_image}
	\end{figure*}

	A schematic description of our system is depicted in Figure \ref{FPGA_pipline_image} (left). We used the Xilinx Kintex 7 chip on the KC705 development board chosen mainly because of the availability of video interfaces. As the output, we used the onboard HDMI output phy, while for the input, we added an external HDMI phy board connected to the main board through an FCM connector.  
	The input frames are received by the board through the HDMI interface in raw Bayer format, 16 bits per pixel with the most significant bytes packaged as the Y channel, and the least significant byte packaged as the 4:2:2 color channels. 
	The input is relayed to Write Agent 0 on the FPGA chip that buffers it in the external dynamic memory. The content of the buffer is brought into the chip by Read Agent 0, which reorders the raster scan order into a stream of $8 \times 8$ patches with configurable amount of overlap. The patches are then fed into a configurable calculator pipeline implementing the reconstruction algorithm detailed in the previous section. The pipeline comprises three configurable stages, one of which is configured as the initial stage (I), another as a middle stage (M), and yet another as the final stage (F), yielding the flow structure of the form \mbox{I $\rightarrow$ $(T-2) \times $ M $\rightarrow$ F}.
	
	The output of the calculator pipeline is produced in 4:2:2 YCbCr format comprising $64$ luma values at $16$ bits per pixel, and additional $32+32=64$ chroma values at $8$ bits per pixel. The luma component undergoes gamma conversion implemented as a lookup table, reducing it to $8$ bits per pixel. The patches are average-pooled (in case of overlap), reordered into raster scan order, and buffered into the dynamic memory by Write Agent 1. Finally, Read Agent 1 conveys the content of the output buffer to HDMI output.

	A schematic block diagram of a calculator stage is depicted in Figure \ref{FPGA_pipline_image} (right). Calculations are performed on vectors in fixed point arithmetics with $16$ precision bits except the multiply-and-accumulate (MACC) block that uses $48$ bit arithmetics internally. To keep a reasonable dynamic range, the data are scaled between various operations by scale factors that were carefully selected to minimize precision loss on a large set of patches from a collection of natural images. Compared to its floating point counterpart, the fixed point implementation produced negligible quality degradation in all our experiments.
	
	Each calculator performs element-wise soft thresholding and the multiplication of the input data by a matrix of size $64 \times 192$ (initial stage, converting the input $64$-dimensional Bayer patch into a set of $192$ coefficients), $192 \times 192$ (middle stage, performing operatons on the coefficients), or $192 \times 128$ (final stage, converting the coefficients into a 4:2:2 YCbCr patch with $64$ luma dimensions and additional $64$ chroma dimensions). This is implemented by using MACC blocks of respective sizes. The parameters of each calculator stage, including threshold values and matrix coefficients, are stored in a local static memory on the FPGA chip. 
	
	Since MACC operations are fully pipelined, they require one clock cycle. The total number of clock cycles it takes a single patch to pass though the chain is given by $64+192\times(T-2)$, where $T-2$ denotes the number of middle stages. There are additional overheads of approximately $100$ cycles per network layer. Due to high resource utilization, we were able to use clock frequency of $125$ MHz only. This results in overall throughput of about $16$ $1920 \times 1080$ frames per second without patch overlap and a $4$ layer network.
	
	%Since MACC operations are fully pipelined, the requires one clock cycle. The total number of clock cycles it takes a single patch to pass though the chain is given by $64+192\times(m+1)$, where $m$ denotes the number of middle stages. There are additional overheads of approximately $100$ cycles per network layer. Due to high resource utilization, we were able to use clock frequency of $125$ MHz only.
	%This results in overall throughput of about $7$ $1962 \times 1080$ (full HD) frames per second using $50\%$ patch overlap and a $4$ layer network. 

	\section{Experimental Evaluation}
	\label{Results}
	\subsection{Synthetic images}
	
	In this experiment, we evaluated the performance of our algorithm on synthetic data of two types. In the first experiment, we used images from the KODAK dataset which were blurred using a PSF simulating typical OOF blur. The same blur kernel was used for the entire image. In the second experiment, we created a synthetic scene with four different depths by using a different kernel in different regions of the image. In all synthetic experiments the reconstruction neural network was trained using $2 \times 10^6$ patches taken from the KODAK training set. 
	%We chose to use an $L_2$ loss for the training process
	%\begin{equation}
	%f(\mbr{x}_n^*,\hat{\mbr{x}_n})=||\mbr{x}_n^*-\hat{\mbr{x}_n}||_2^2 ,
	%\end{equation} 
	The network with $T=8$ layers was converted to fixed-point arithmetics as described in Section \ref{sec:FPGA}.
	
	\paragraph{Single Depth.}
	
\begin{figure}[h!]
		\centering	
		\begin{tabular}{ c }
			
			\includegraphics[width = 0.75\textwidth]{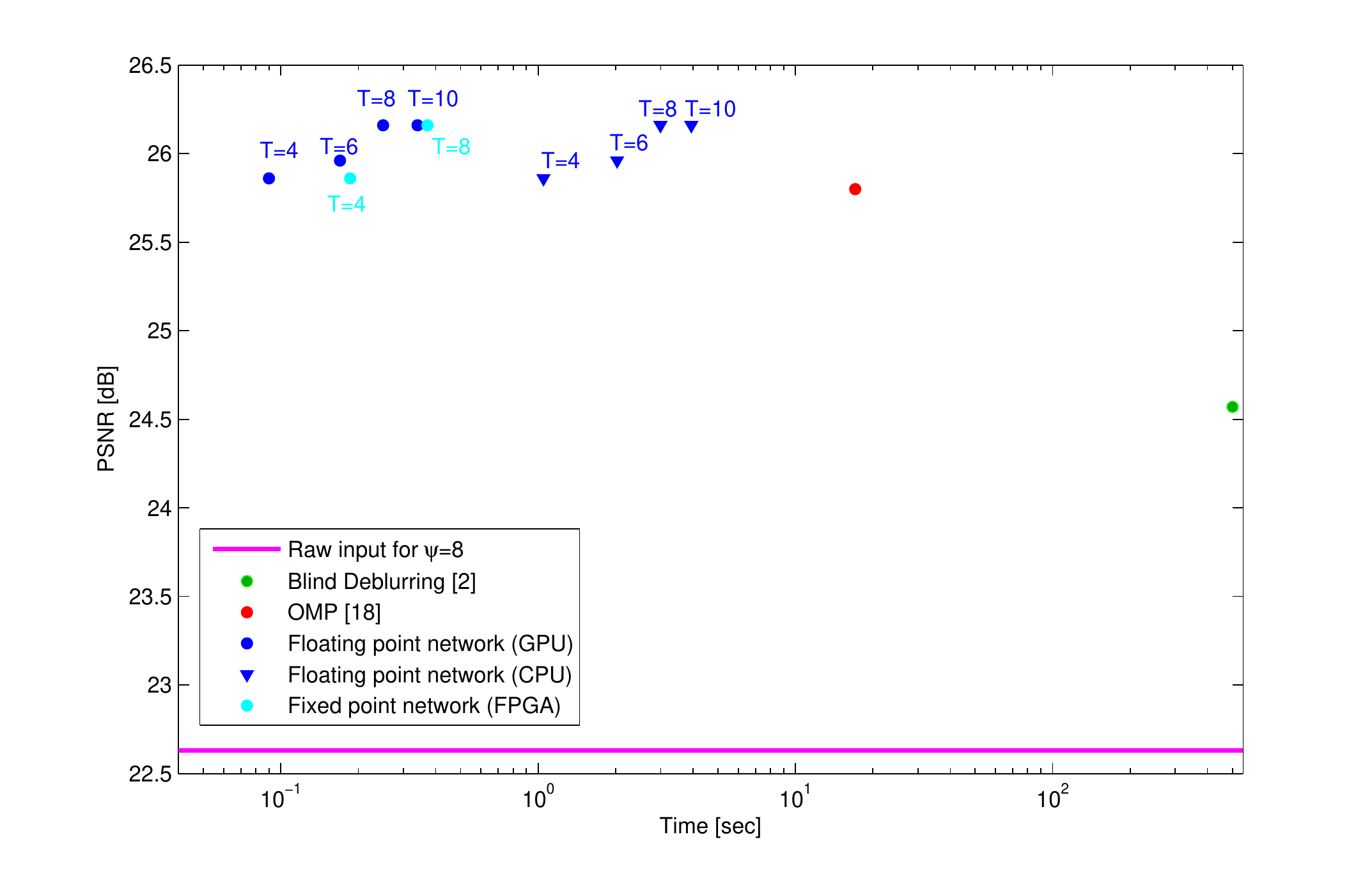} \\
			
		\end{tabular}   
		\caption{\small \red{\textbf{Comparison of average PSNR and run time on KODAK images}. Presented are average values over all test images in the KODAK dataset with simulated blur using $\psi=8$ (magenta) reconstructed with the different algorithms. Top left corner represents high reconstruction quality and low runtime. One can see that our algorithm (blue and cyan) achieves superior reconstruction quality in comparison to the blind deblurring algorithm of \cite{krishnan2011blind} (green) and to the OMP method of \cite{haim2014multi} (red). In addition, it is evident that increasing the number of layers (denoted by $T$ above the blue and cyan points) improves the performance at an additional computational cost. Our fixed point FPGA implementation (cyan) offers the same reconstruction quality as its floating point (blue) counterpart executed on the CPU while being an order of magnitude faster. Remarkably, the FPGA implementation run-time is in the same range as the GPU floating point implementation. All patch based algorithms were run using a patch stride of $2$ pixels. Executing times were measured on an Intel Xeon E5-2650 2GHz CPU and a Tesla C2075 1.15GHz GPU. Our fixed-point implementation was executed on a 100MHz Xilinx Kintex 7 FPGA.   }}
		\label{table_synthetic_PSNR}
	\end{figure}

	Images taken from the KODAK dataset were used for the evaluation. Each image was convolved with a PSF corresponding to $\psi=8$ and mosaiced to simulate the input to the system. The algorithms compared were \red{ our OMP implementation of \cite{haim2014multi}} with a $k=192$ atom dictionary trained using $k$-SVD and \red{our} reconstruction neural network with a varying number of layers $T$. As a reference, we compared our algorithms to the blind deblurring algorithm from \cite{krishnan2011blind} following MATLAB default demosaicing algorithm. As the PSF varies across color channels, the algorithm was run separately on each channel. Image reconstruction quality in terms of average PSNR and execution times are presented in Figure \ref{table_synthetic_PSNR}. It is evident that the highest PSNR is achieved by the neural network with $T=10$; restricting to $T=8$ layers performs twice as fast yielding almost the same average PSNR score. Interestingly, the neural network achieves better reconstruction quality compared to the greedy OMP algorithm \red{\cite{haim2014multi}}, which we attribute to the effect of supervised training.  All sparse prior-based algorithms outperform the blind deconvolution algorithm \cite{krishnan2011blind} by over $1$ dB PSNR. Comparing the execution times of the algorithms on a standard CPU shows that the OMP algoithm is about $30$ times faster than \cite{krishnan2011blind}, while all neural network implementations are more than $15$ times faster than OMP and $500$ times fatser than \cite{krishnan2011blind} with superior reconstruction quality.
	\vspace{-0.5cm}
	
	\begin{figure}[h!]
		\centering	
		\begin{tabular}{ c c c c c}
			\includegraphics[width = 0.17\textwidth]{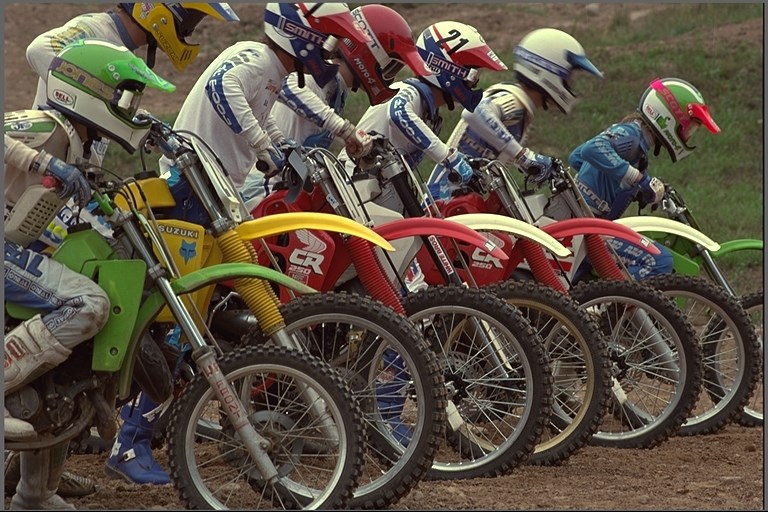} &
			\includegraphics[width = 0.17\textwidth]{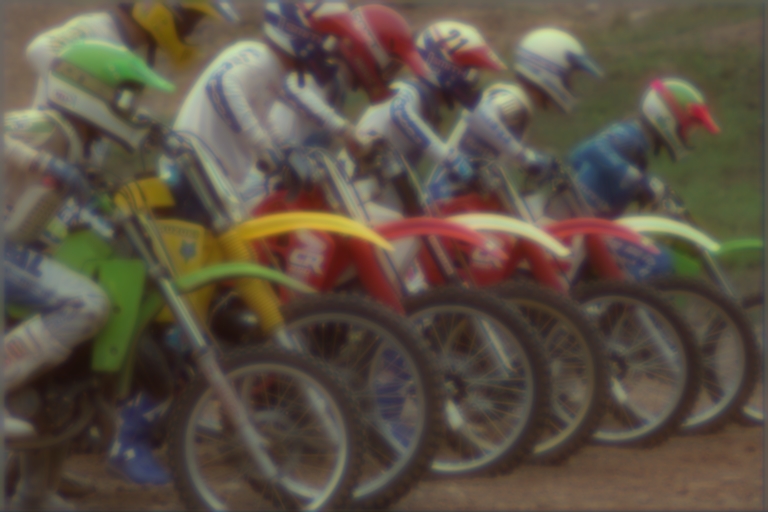} &
			\includegraphics[width = 0.17\textwidth]{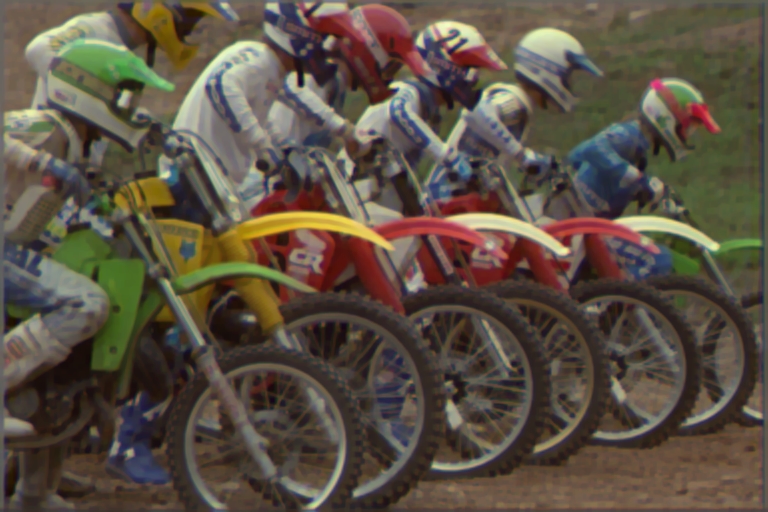} &
			\includegraphics[width = 0.17\textwidth]{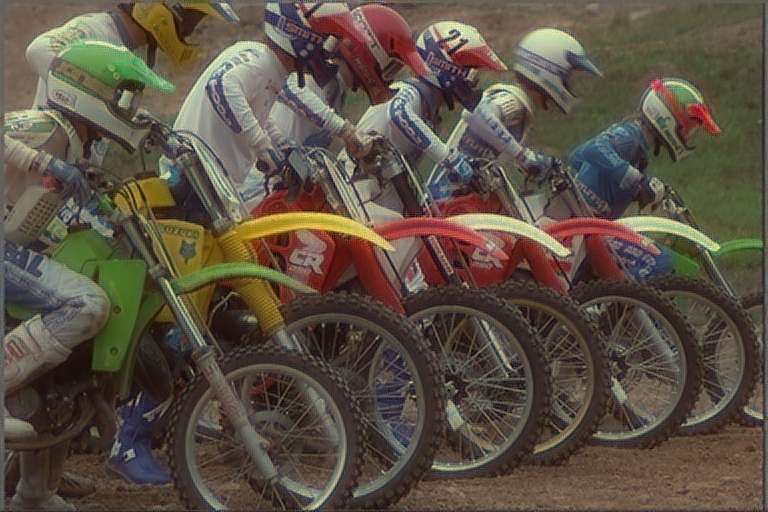} &
			\includegraphics[width = 0.17\textwidth]{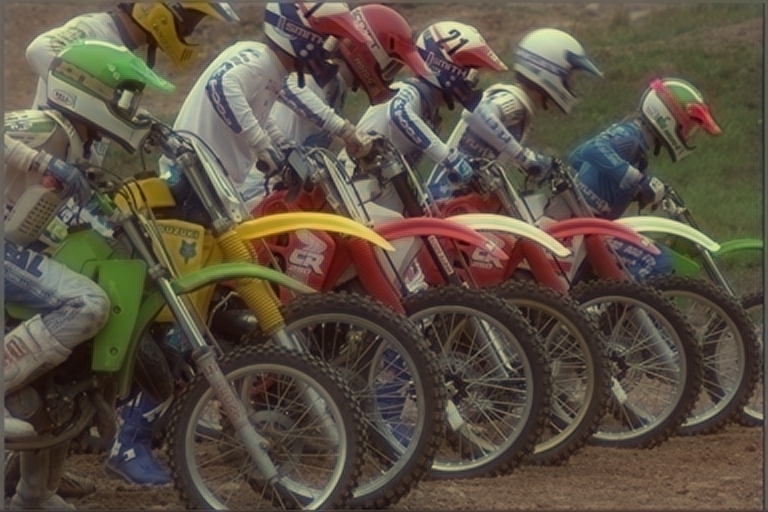} \\
			
			\includegraphics[width = 0.17\textwidth]{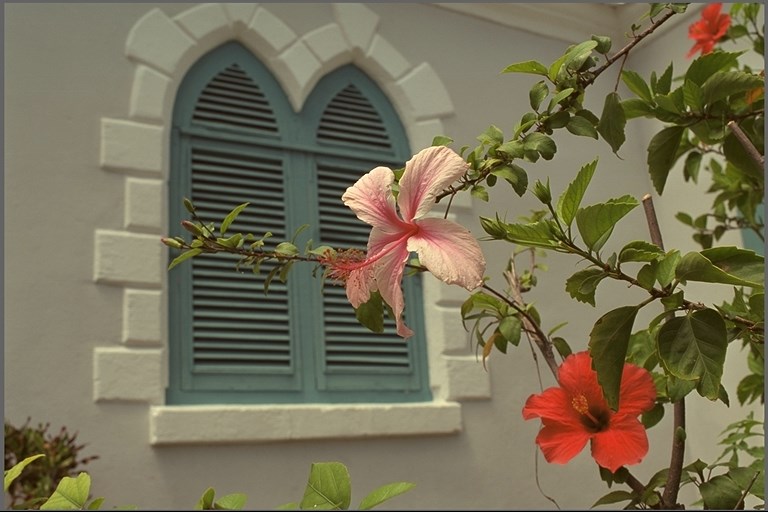} &
			\includegraphics[width = 0.17\textwidth]{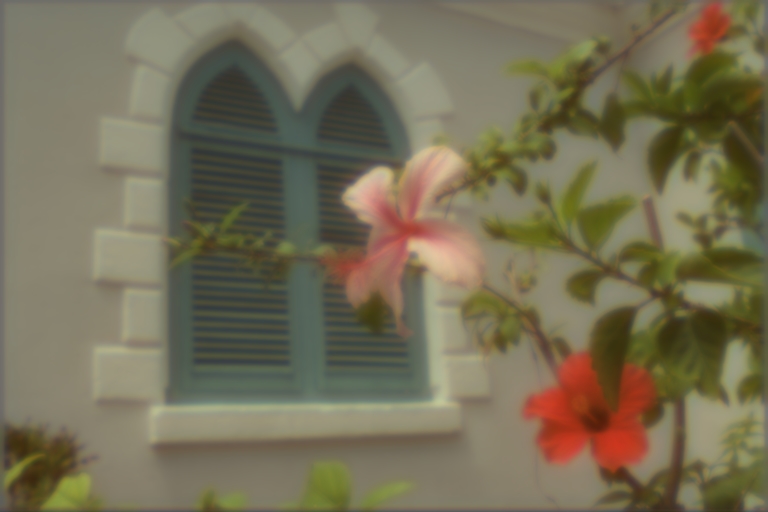} &
			\includegraphics[width = 0.17\textwidth]{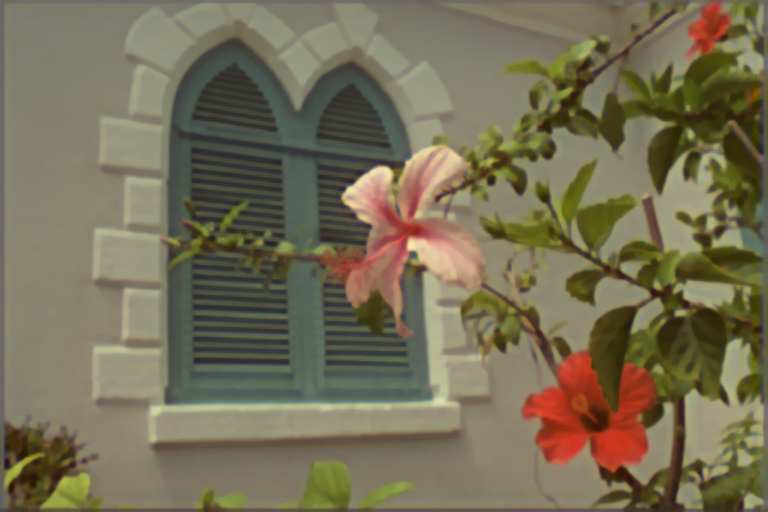} &
			\includegraphics[width = 0.17\textwidth]{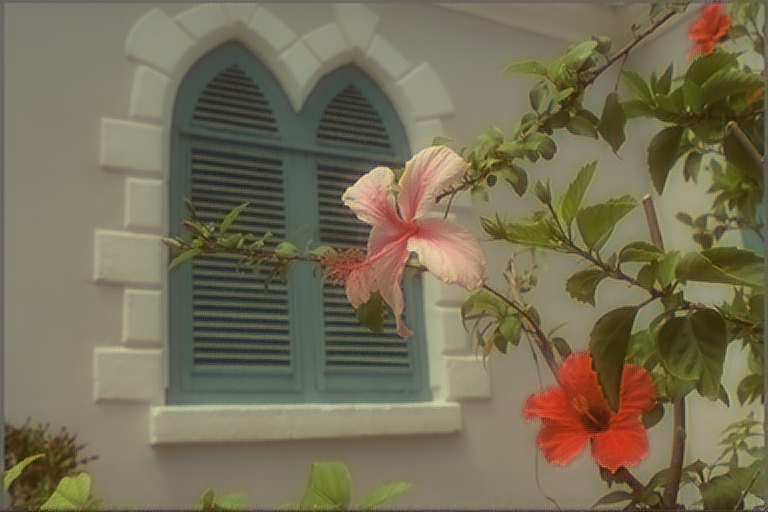} &
			\includegraphics[width = 0.17\textwidth]{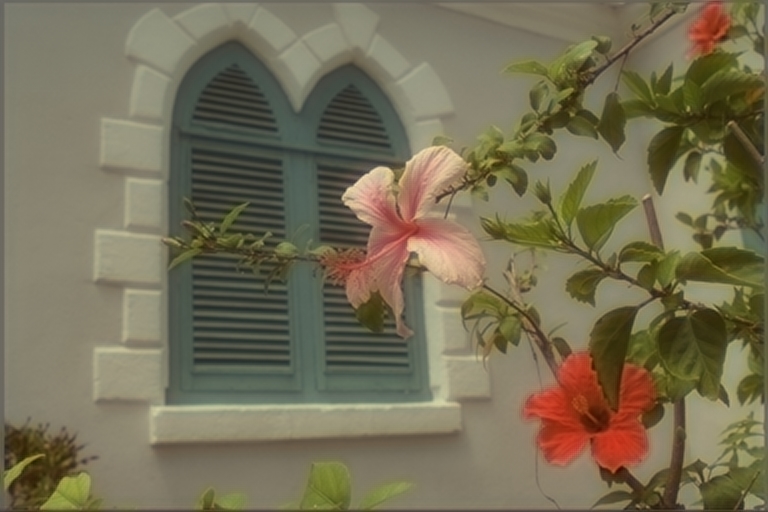} \\
			
			\includegraphics[width = 0.17\textwidth]{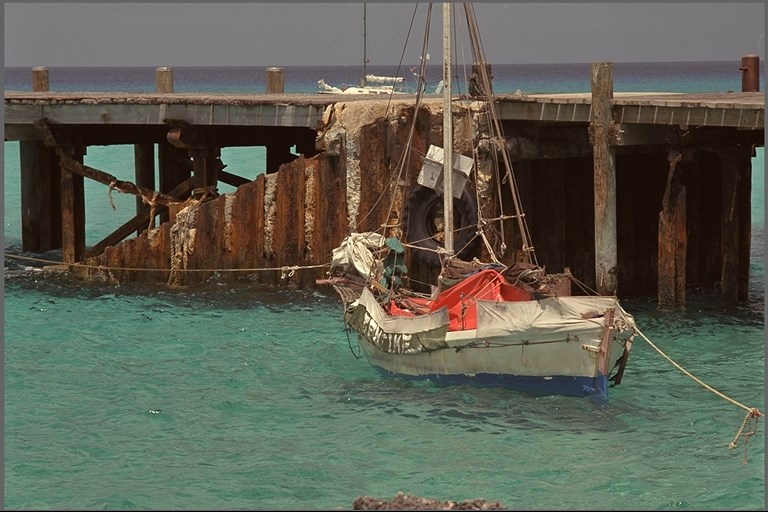} &
			\includegraphics[width = 0.17\textwidth]{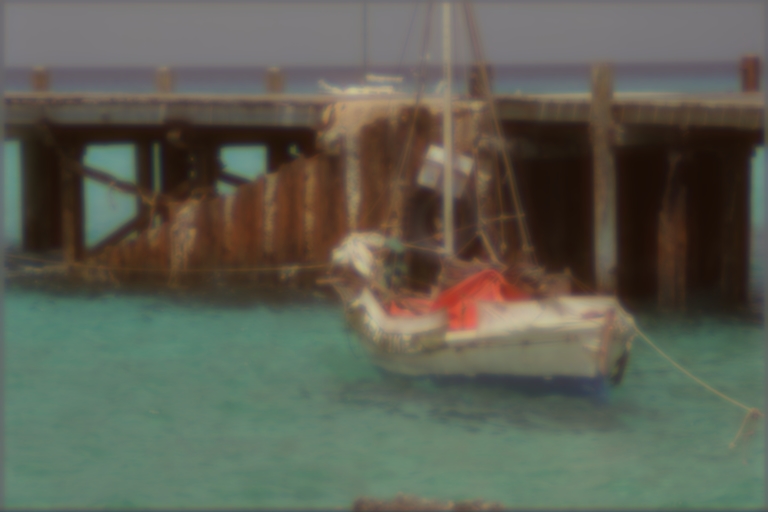} &
			\includegraphics[width = 0.17\textwidth]{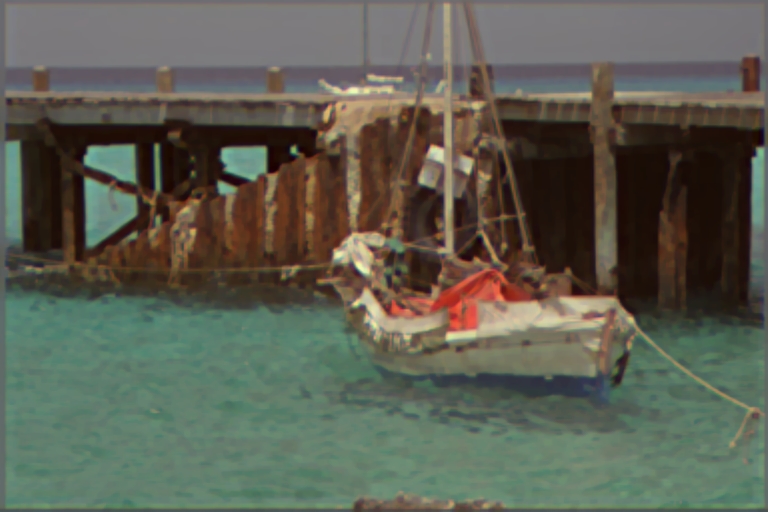} &
			\includegraphics[width = 0.17\textwidth]{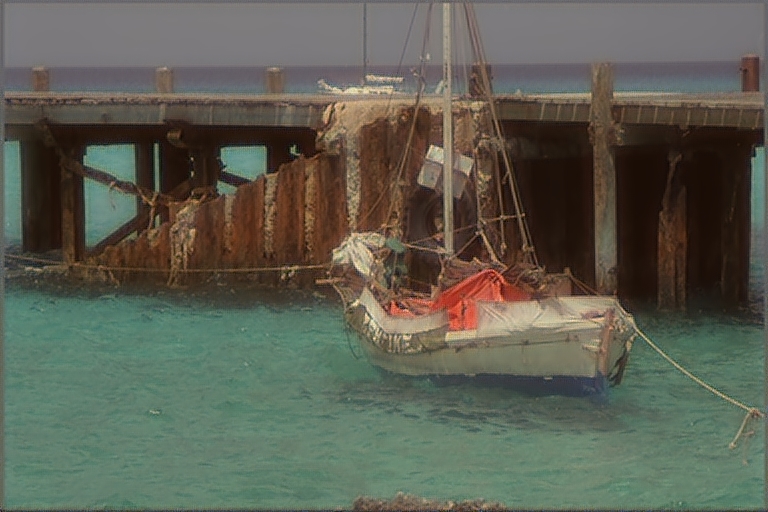} &
			\includegraphics[width = 0.17\textwidth]{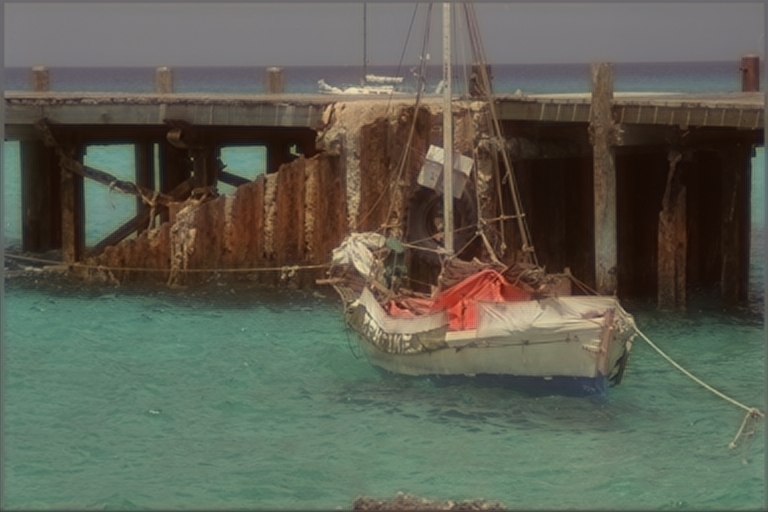} \\
			
			\includegraphics[width = 0.17\textwidth]{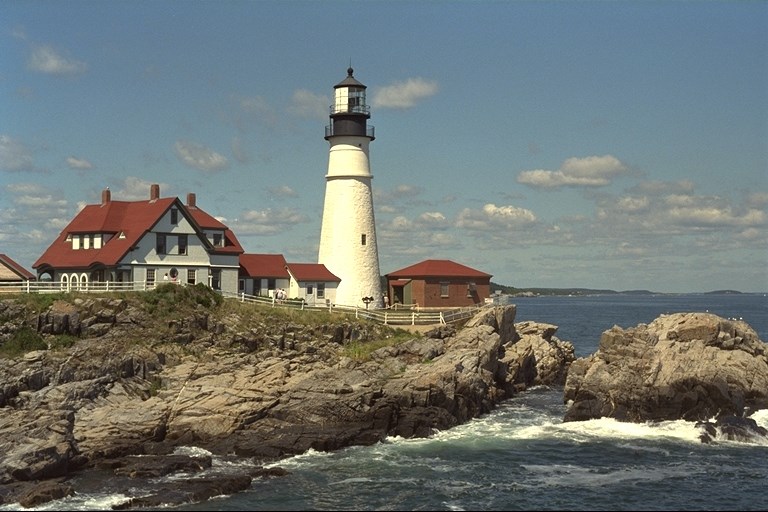} &
			\includegraphics[width = 0.17\textwidth]{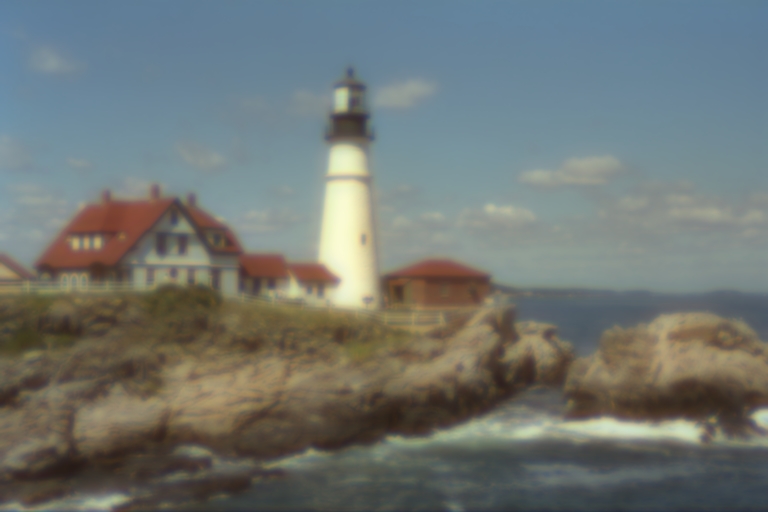} &
			\includegraphics[width = 0.17\textwidth]{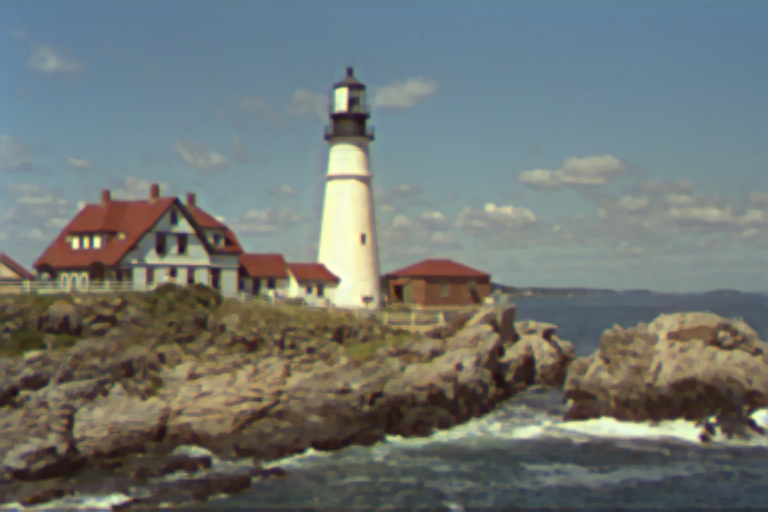} &
			\includegraphics[width = 0.17\textwidth]{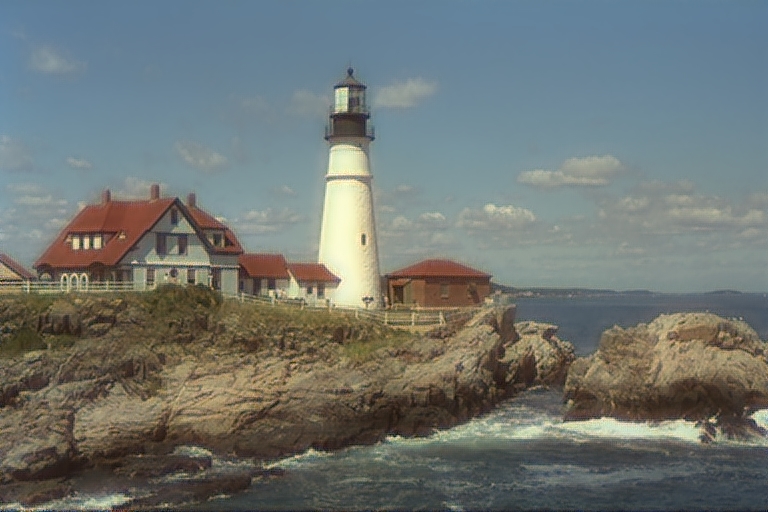} &
			\includegraphics[width = 0.17\textwidth]{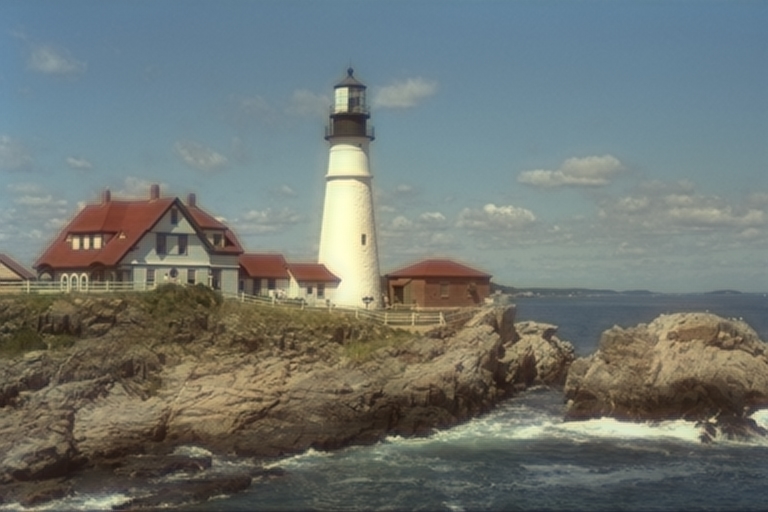} \\
			
			\includegraphics[width = 0.17\textwidth]{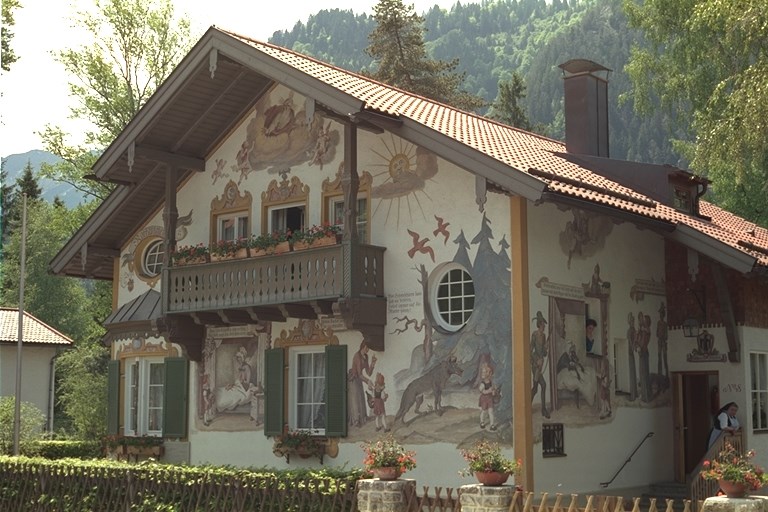} &
			\includegraphics[width = 0.17\textwidth]{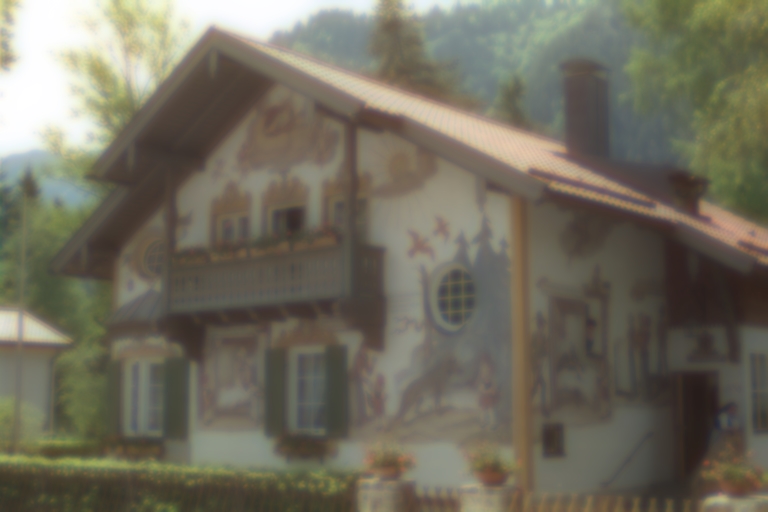} &
			\includegraphics[width = 0.17\textwidth]{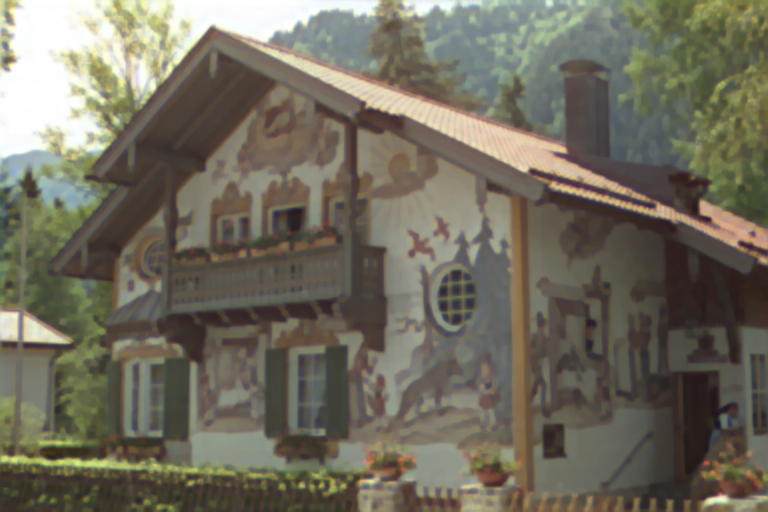} &
			\includegraphics[width = 0.17\textwidth]{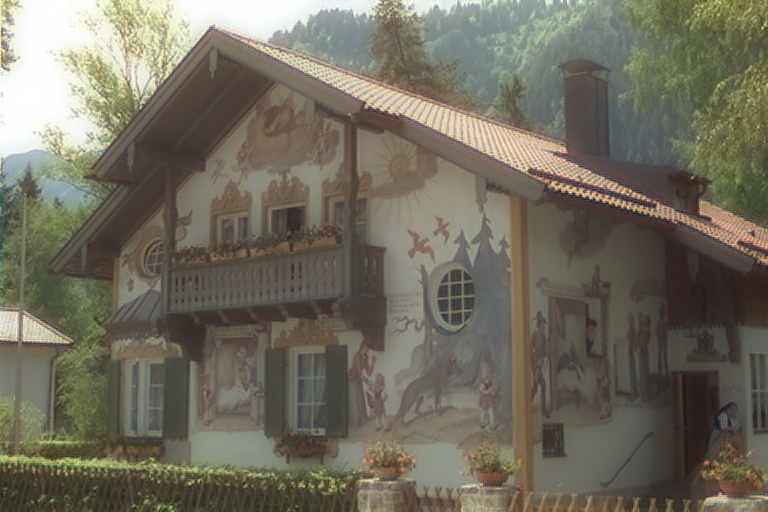} &
			\includegraphics[width = 0.17\textwidth]{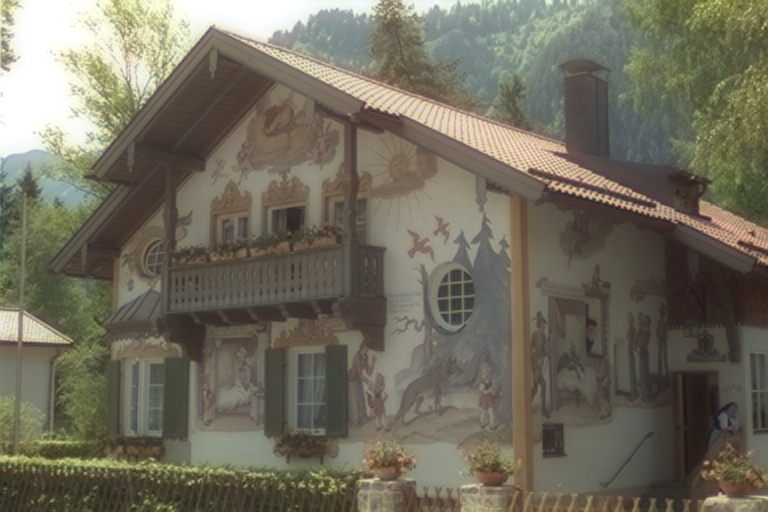} \\
			
			\red{Ground truth} & \red{Blurred} & \cite{krishnan2011blind} & \cite{haim2014multi} & Our FPGA
			
		\end{tabular}   
		\caption{\small \red{ \textbf{Synthetic image reconstruction.} Columns from left to right: original scene, blurred scene (using a value of $\psi=8$), blind deblurring using \cite{krishnan2011blind}, our implementation of the OMP reconstruction of \cite{haim2014multi} and our FPGA fixed-point neural network with $T=8$ layers.}}
		\label{table_synthetic_images}
	\end{figure}
	\noindent \red{In a GPU implementation, our neural network is more than $2,000$ times faster than \cite{krishnan2011blind} and $70$ times faster than the OMP method of \cite{haim2014multi}. Furthermore, our fixed point FPGA implementation offers the same reconstruction quality as its floating point counterpart and is an order of magnitude faster than the CPU implementation. Remarkably, the FPGA implementation run-time is in the same range as the floating point GPU implementation, while running at an order of magnitude slower clock and requiring over two orders of magnitude less power. }
	Ground truth images and the reconstruction results by the different algorithms are presented in Figure \ref{table_synthetic_images}. Qualitatively, one can observe that the fixed-point neural network algorithm with $T=8$ layers outperforms all other algorithms presented in the figure. Our experiments show that the output of fixed-point network with $T=4$ is almost imperceavably inferior to the former case, while requiring significantly lower complexity.
	
	\paragraph{Multiple Depths.}
	
\begin{figure}[h]
		\centering
		\begin{tabular}{ c }
			
			{
				\begin{tabular}{ c c c }
					\multicolumn{1}{ c }{\includegraphics[width=0.32\textwidth]{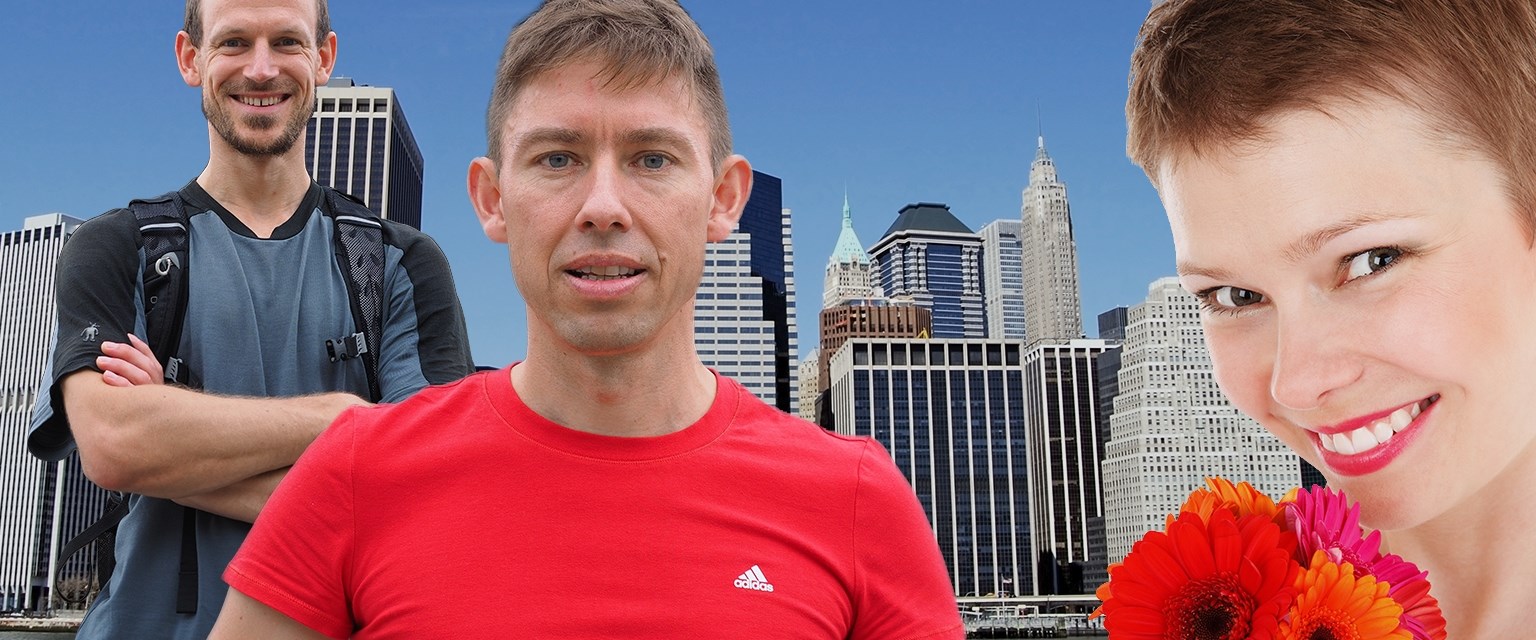}}  &				
					\multicolumn{1}{ c }{\includegraphics[width=0.32\textwidth]{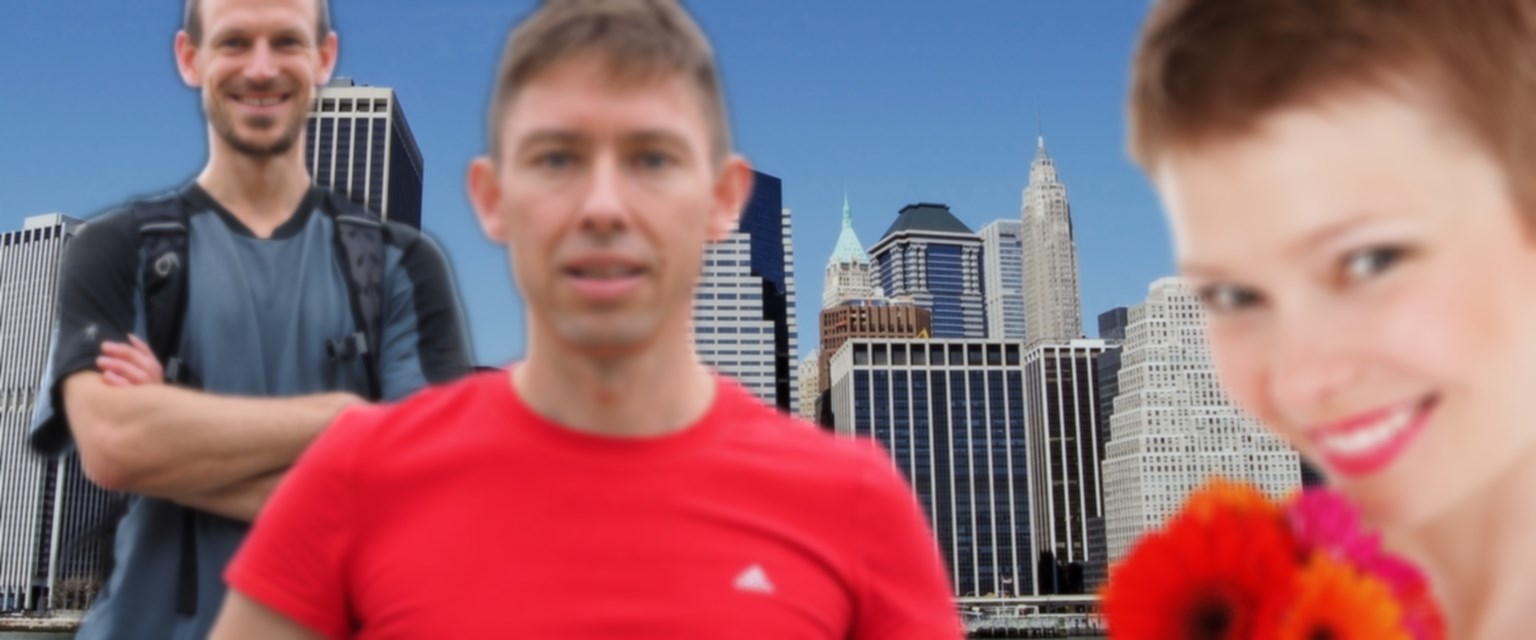}}  &
					\multicolumn{1}{ c }{\includegraphics[width=0.32\textwidth]{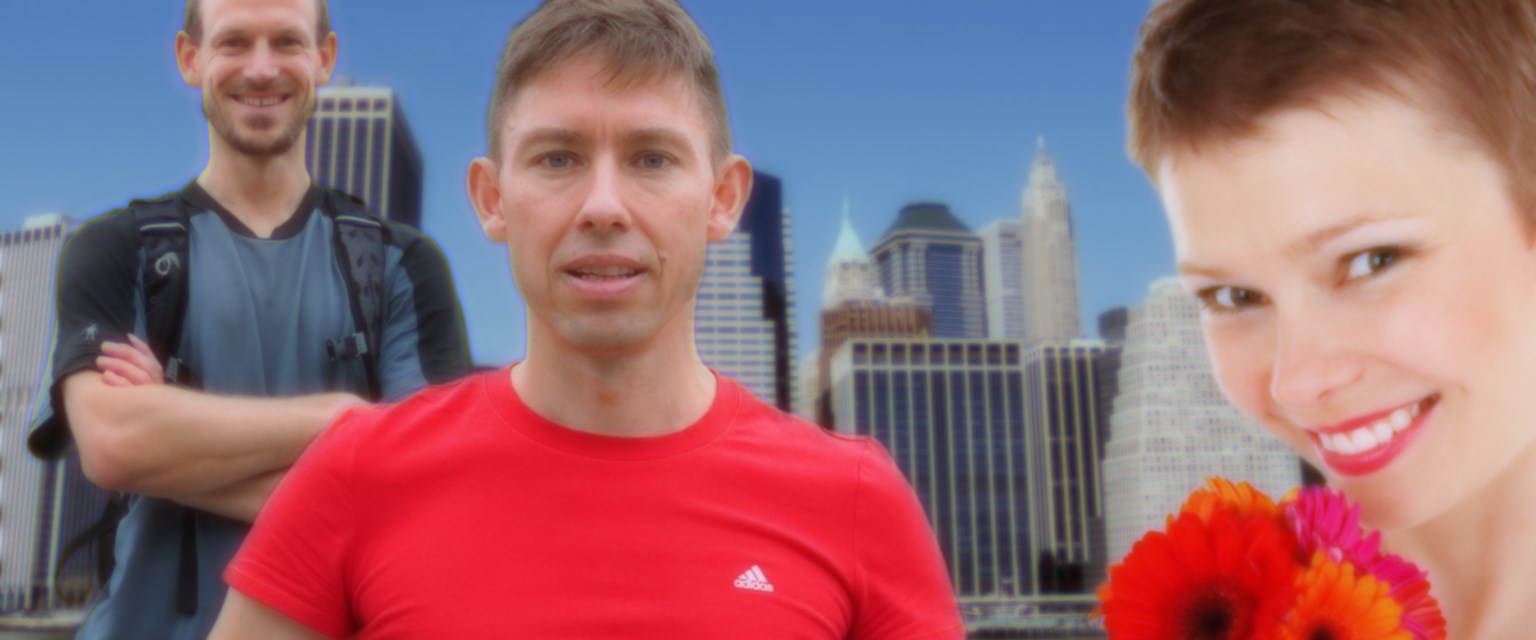}}  \\
					\multicolumn{1}{ c }{Ground truth} &
					\multicolumn{1}{ c }{Conventional imaging} &
					\multicolumn{1}{ c }{Imaging with phase coding} \\
				\end{tabular}	
			}\vspace{0.5cm}\\{
			\begin{tabular}{ c c c c c}	
				
				\multicolumn{1}{ c }{\includegraphics[width=0.19\textwidth]{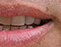}} &
				\multicolumn{1}{ c }{\includegraphics[width=0.19\textwidth]{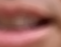}} &
				\multicolumn{1}{ c }{\includegraphics[width=0.19\textwidth]{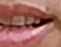}} &			
				\multicolumn{1}{ c }{\includegraphics[width=0.19\textwidth]{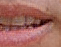}} &			
				\multicolumn{1}{ c }{\includegraphics[width=0.19\textwidth]{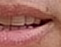}} \\
				
				\multicolumn{1}{ c }{\includegraphics[width=0.19\textwidth]{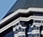}} &
				\multicolumn{1}{ c }{\includegraphics[width=0.19\textwidth]{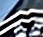}} & 
				\multicolumn{1}{ c }{\includegraphics[width=0.19\textwidth]{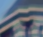}} & 
				\multicolumn{1}{ c }{\includegraphics[width=0.19\textwidth]{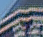}} & 
				\multicolumn{1}{ c }{\includegraphics[width=0.19\textwidth]{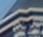}} \\
				
				\multicolumn{1}{ c }{\includegraphics[width=0.19\textwidth]{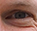}} &
				\multicolumn{1}{ c }{\includegraphics[width=0.19\textwidth]{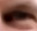}} &		
				\multicolumn{1}{ c }{\includegraphics[width=0.19\textwidth]{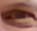}} &
				\multicolumn{1}{ c }{\includegraphics[width=0.19\textwidth]{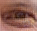}} &					
				\multicolumn{1}{ c }{\includegraphics[width=0.19\textwidth]{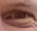}} \\	
				
				\multicolumn{1}{ c }{\footnotesize Ground truth} &
				\multicolumn{1}{ c }{\footnotesize Clear + \cite{krishnan2011blind}} &
				\multicolumn{1}{ c }{\footnotesize Phase + \cite{krishnan2011blind}} &
				\multicolumn{1}{ c }{\footnotesize Phase + \cite{haim2014multi}}&
				\multicolumn{1}{ c }{\footnotesize Phase + our} \\			
				
				\multicolumn{1}{ c }{} &
				\multicolumn{1}{ c }{\footnotesize $17.95$ $\rm{dB}$} &
				\multicolumn{1}{ c }{\footnotesize $22.75$ $\rm{dB}$} &
				\multicolumn{1}{ c }{\footnotesize $24.78$ $\rm{dB}$}&
				\multicolumn{1}{ c }{\footnotesize $24.83$ $\rm{dB}$} \\
				
			\end{tabular}  } 
		\end{tabular}
		\caption{\red{\small \textbf{Synthetic Multiple Depth.} The top row shows from left to right: the original synthetic image used for simulating imaging effect on multi depth scene where each object was blurred according to its distance from the camera (the focus point was on the background buildings); Imaging results for conventional clear aperture imaging and imaging with a phase-coded aperture. The three bottom rows present from left to right  zoomed-in snippets as follows: Ground truth, blind deblurring proposed by \cite{krishnan2011blind} applied on the clear aperture image acquired using a conventional imaging system,  blind deblurring of the phase-coded aperture image acquired by our optical system using \cite{krishnan2011blind}, OMP \cite{haim2014multi} and our neural network. PSNR values are reported below the images. The full sized images are available in the supplementary material. 
			}}	
			\label{table_synthetic_2p5D_images}
		\end{figure}

	As indicated in the previous section, for natural depth scenes one cannot assume that the image is blurred by a single blurring kernel. Our reconstruction process analyzes small patches rather than the entire image; therefore, the process is applied to every region inside the image independently of the other regions, allowing our algorithm to treat the input as if it had a single blur kernel.
	\red{To demonstrate the process, we simulated a "$2.5\mathrm{D}$" scene with four objects each located at a different constant distance from the camera as shown in Figure \ref{table_synthetic_2p5D_images} (top left). The middle image in the top row shows the simulation of the scene acquired using a conventional camera focused on the background. As expected, other objects in the scene are increasingly blurred with the increase of their distance from the focus point. The rightmost image in the top row presents a simulated acquisition through a phase-coded aperture. The three bottom rows of Figure \ref{table_synthetic_2p5D_images} present zoomed-in snippets  of the ground truth compared against reconstruction results using our proposed method, the blind deblurring algorithms \cite{krishnan2011blind} and \cite{haim2014multi} applied to the clear and to the phase-coded aperture images. }

	\red{
		Observe that by using our method all objects were restored without noticeable artifacts as opposed to the blind restoration of \cite{krishnan2011blind} applied on both aperture types. Quantitatively, our proposed method outperforms all other methods. While the improvement in terms of PSNR over the greedy OMP algorithm proposed by \cite{haim2014multi} is minor, a noticeable qualitative improvement is apparent in the zoomed-in snippets. 
	}

	\subsection{Real optical system}
\begin{figure}[h]
		\centering
		\begin{tabular}{ c c }
			\includegraphics[width = 0.34\textwidth]{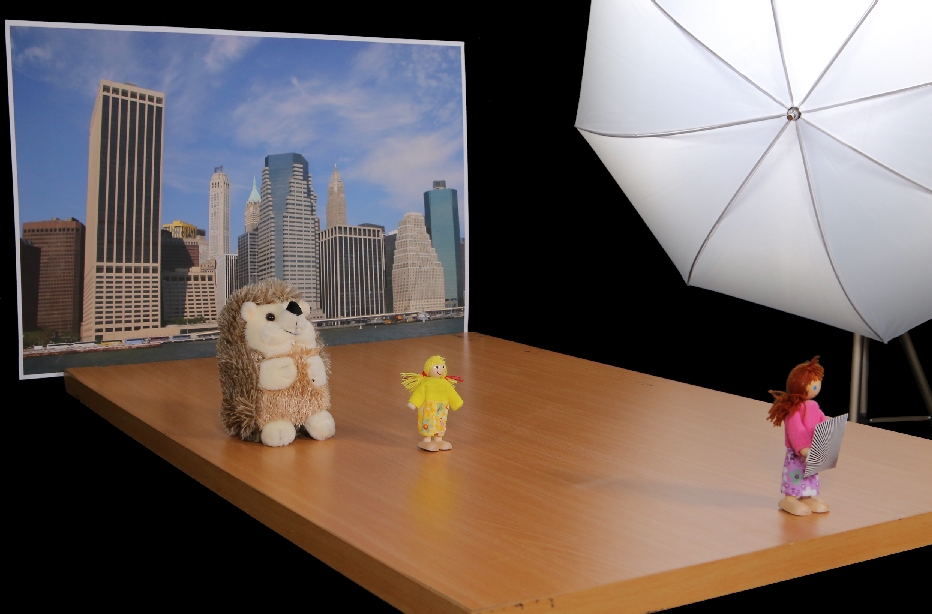} &
			\includegraphics[width = 0.34\textwidth]{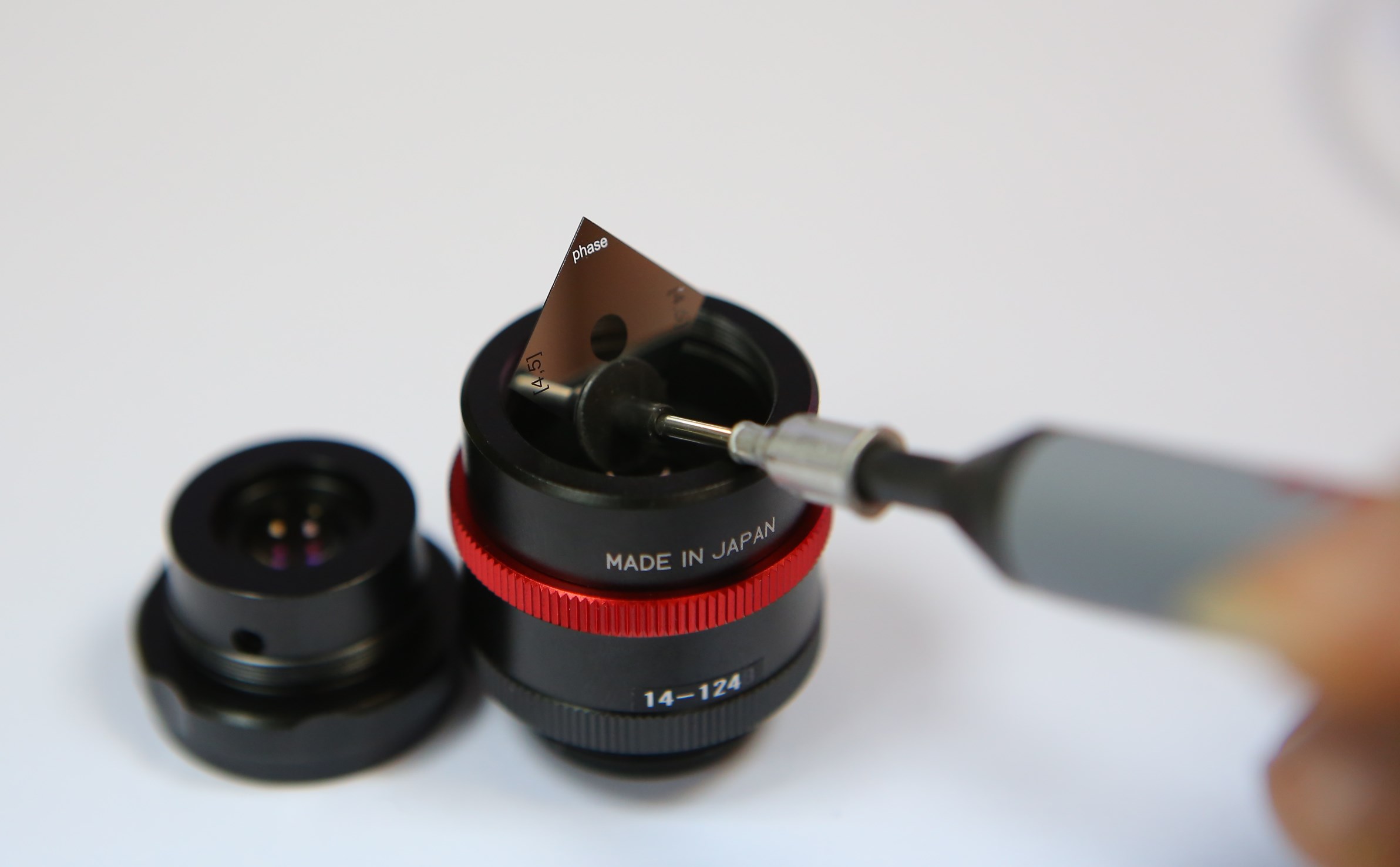}  \\
			(a) & (b)
			
		\end{tabular}   \\
		\caption{\small \textbf{System setup.} (a) table setup of the scene presented in Figure \ref{table_DSLR_images}; (b) $16$mm lens with the phase mask (red arrow) inserted into its aperture.}
		\label{system_setup}
	\end{figure}

	We built a table-top experimental system consisting of a conventional CMOS camera and a 16mm C-mount commercial lens, into which the phase mask was inserted (Figure \ref{system_setup},b). 
	\vspace*{-0.5cm}
	
\begin{figure}[h!]
		\centering
		%\vspace*{-4cm} 
		\begin{tabular}{ c c c c c c}			
			
			\includegraphics[width=0.15\textwidth]{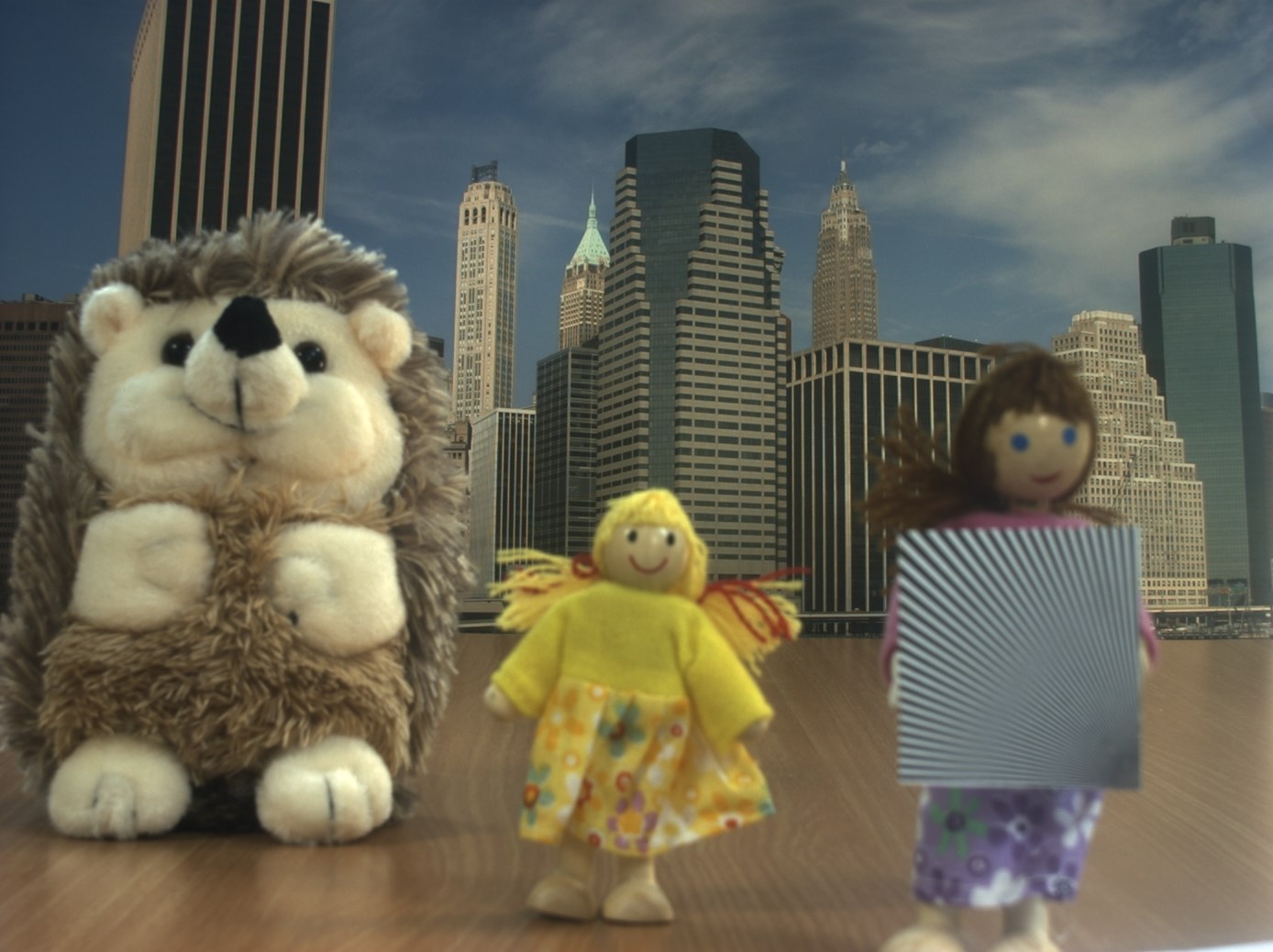} &
			\includegraphics[width=0.15\textwidth]{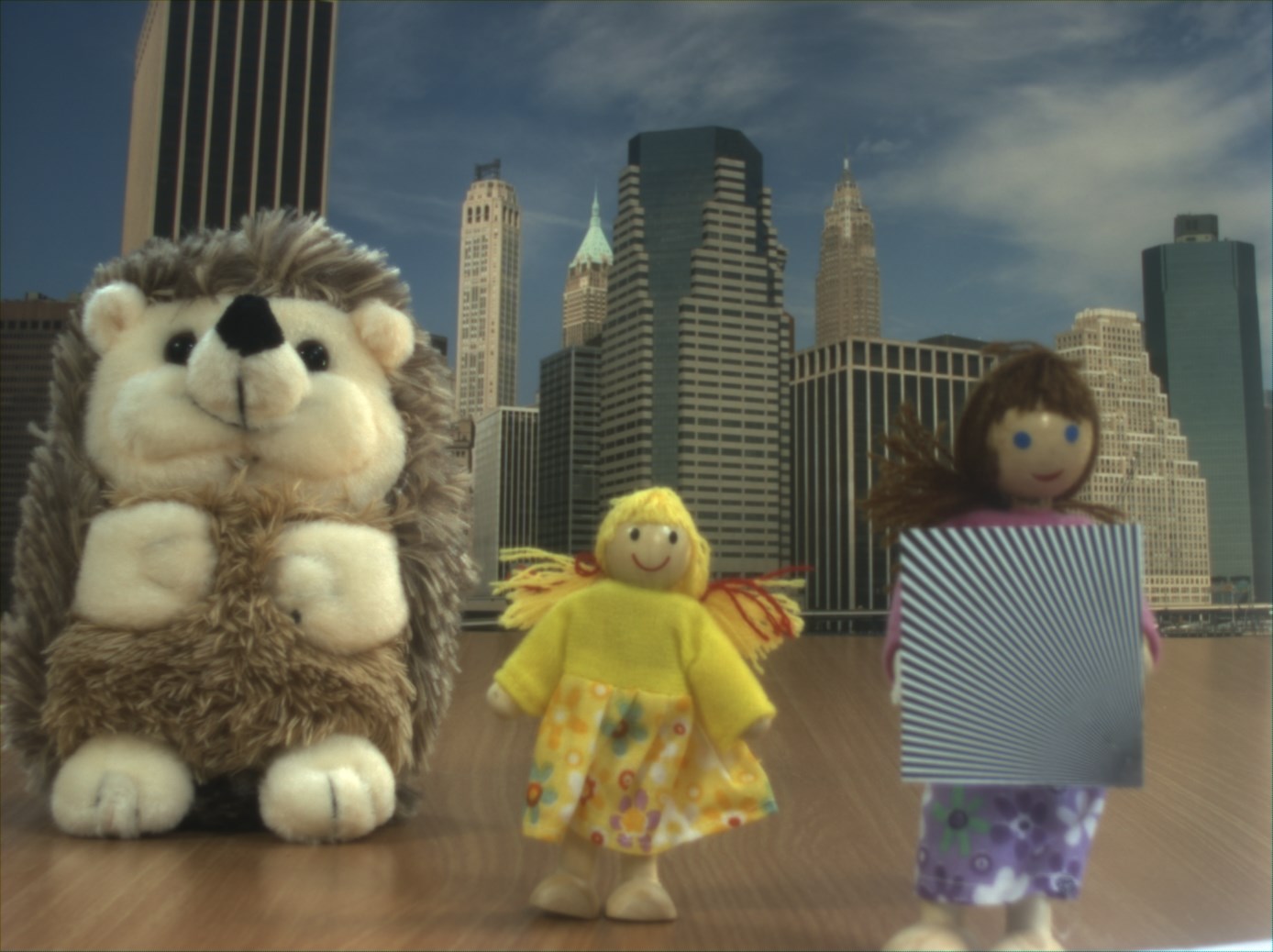} &
			\includegraphics[width=0.15\textwidth]{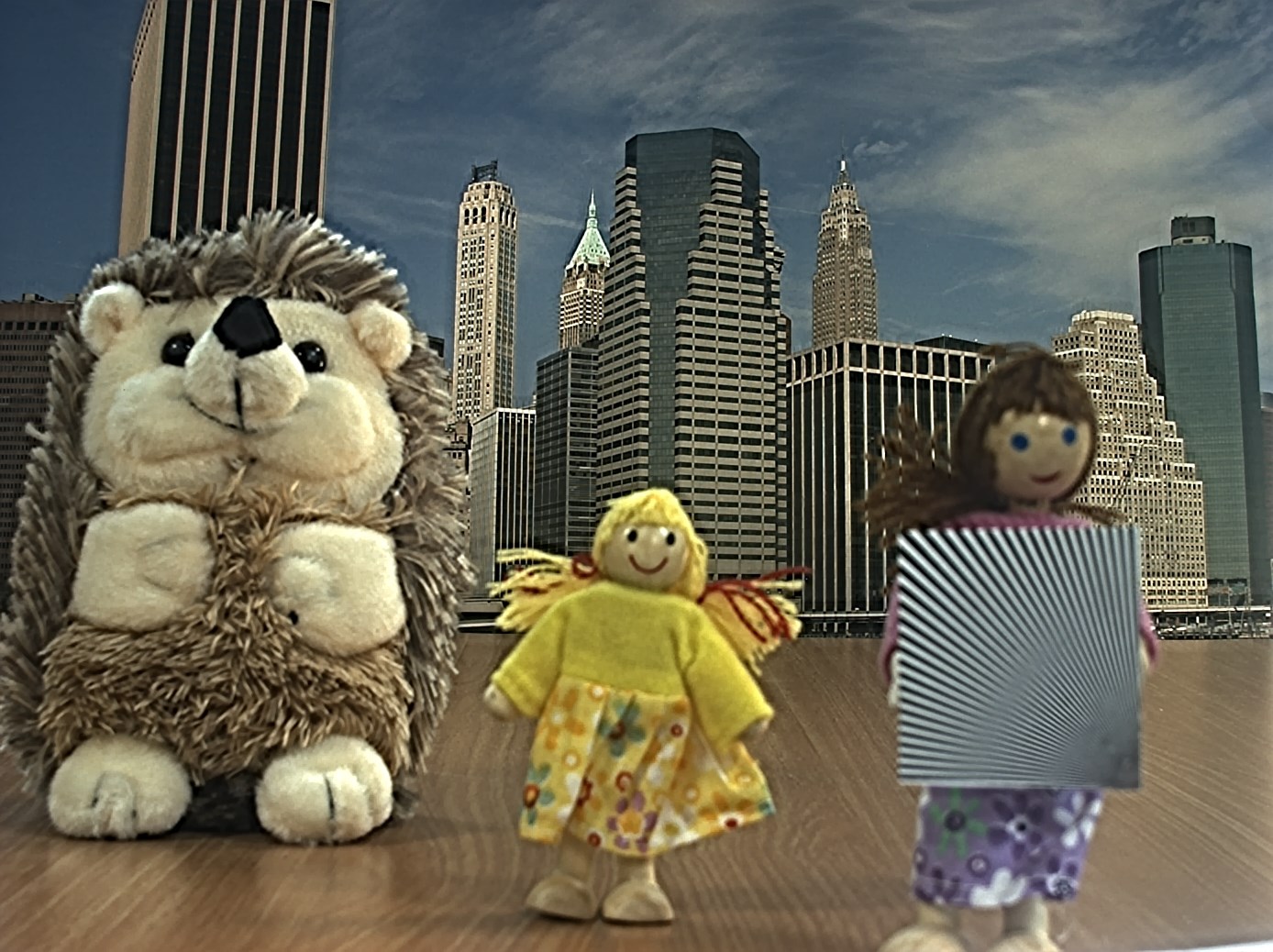} &
			\includegraphics[width=0.15\textwidth]{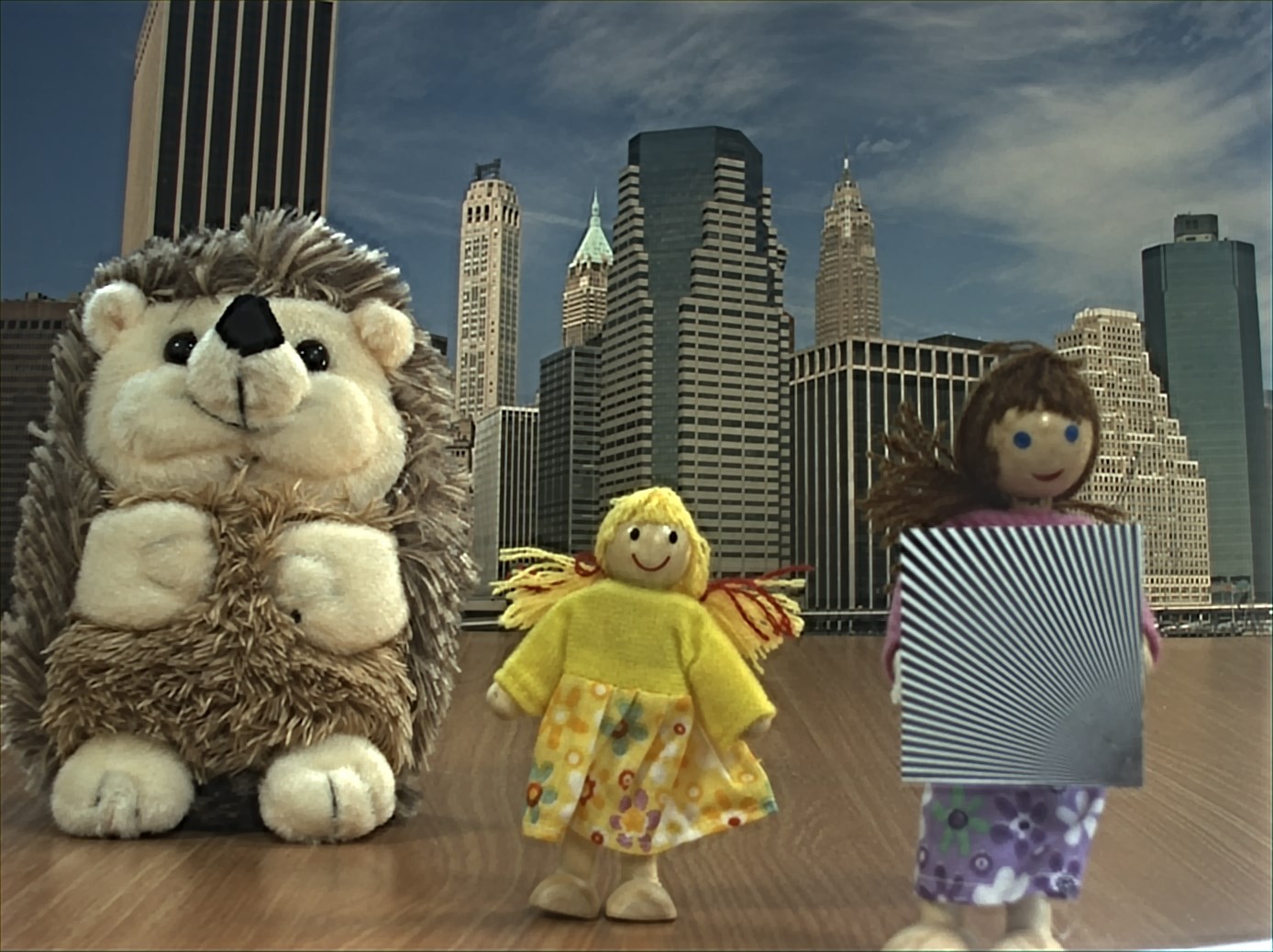} &
			\includegraphics[width=0.15\textwidth]{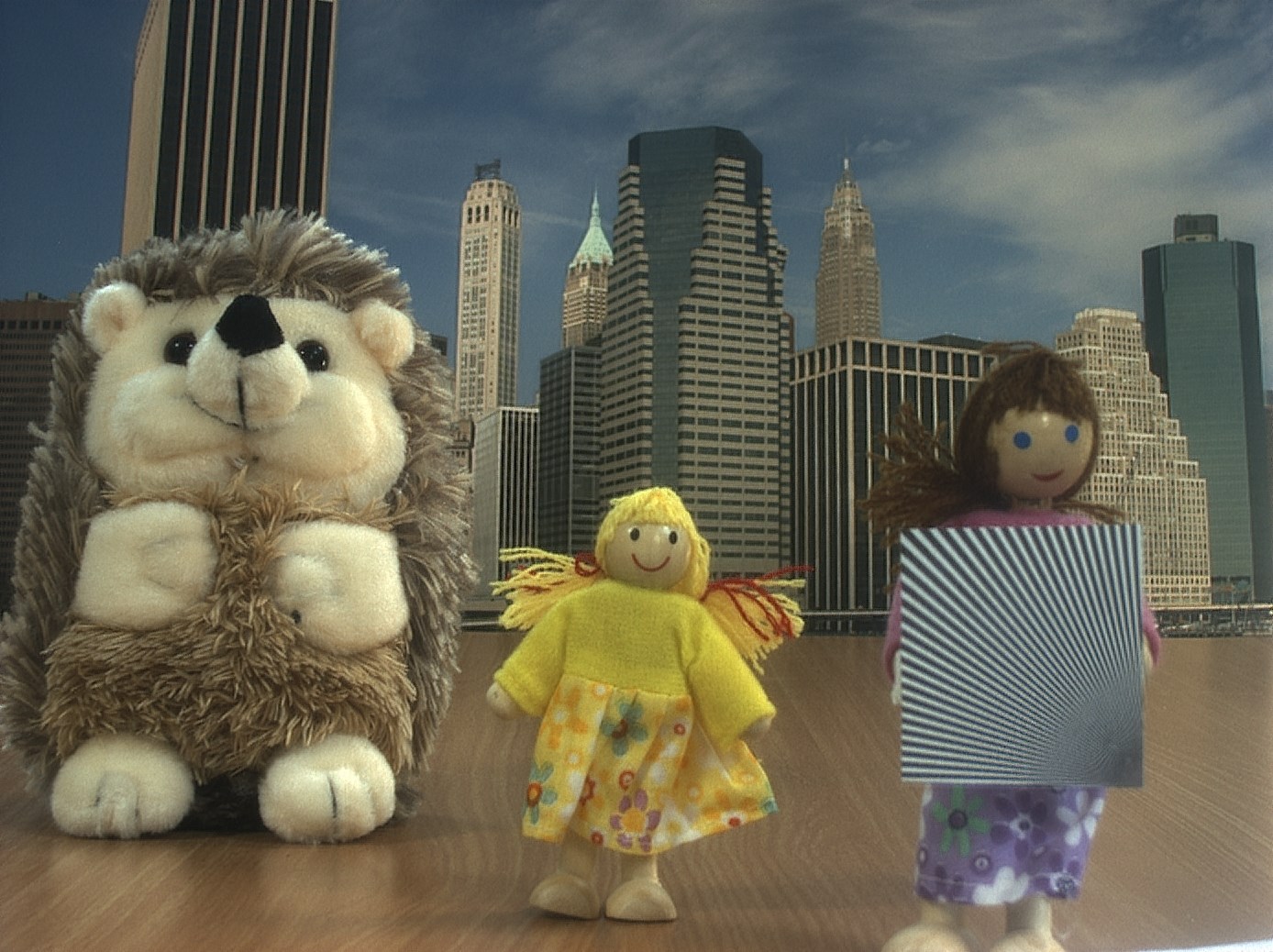} &
			\includegraphics[width=0.15\textwidth]{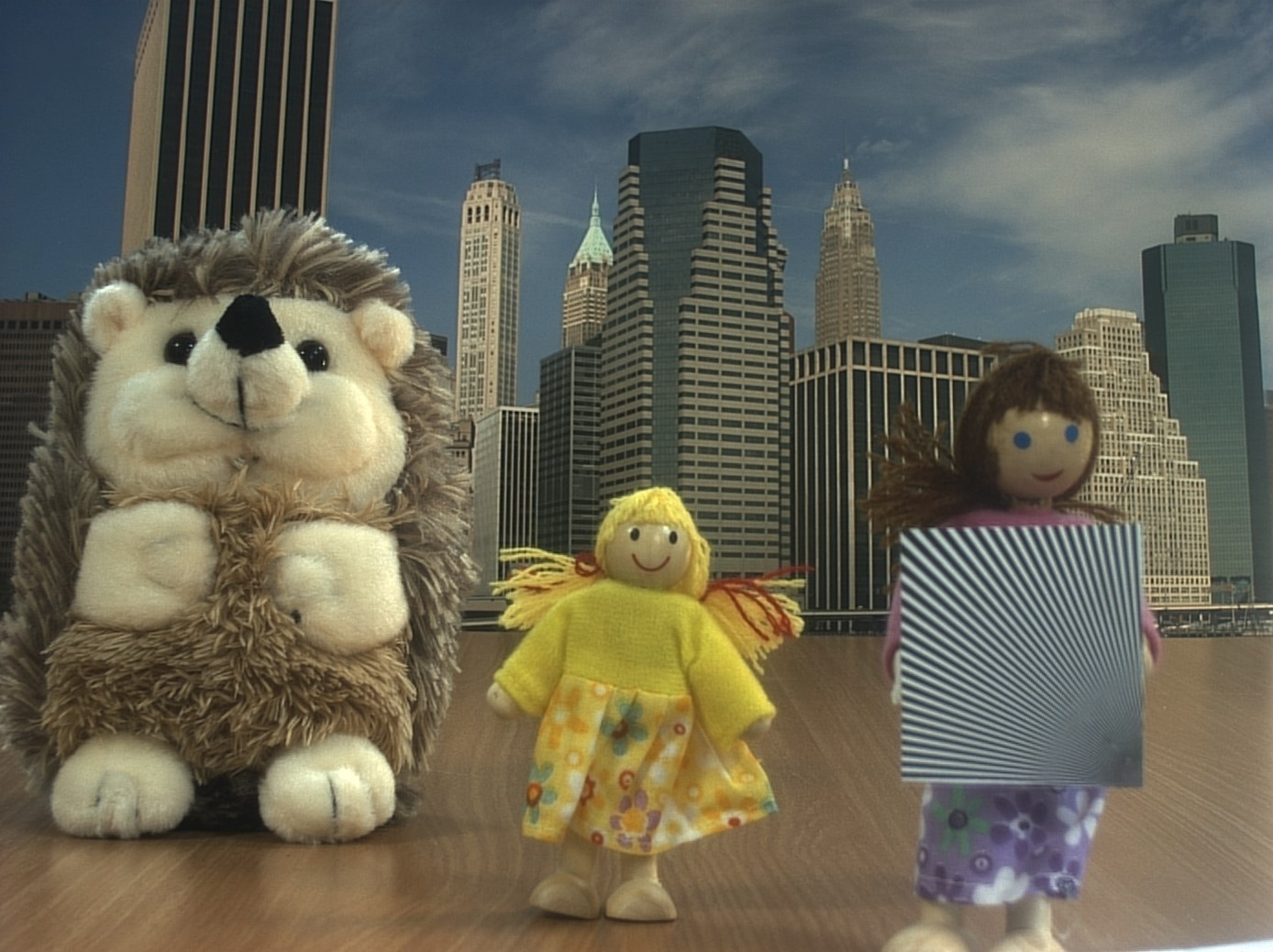} \\		
			
			\includegraphics[width=0.15\textwidth]{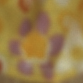} &
			\includegraphics[width=0.15\textwidth]{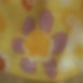} &
			\includegraphics[width=0.15\textwidth]{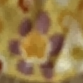} &
			\includegraphics[width=0.15\textwidth]{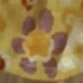} &	
			\includegraphics[width=0.15\textwidth]{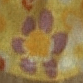} &
			\includegraphics[width=0.15\textwidth]{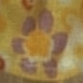} \\
			
			\includegraphics[width=0.15\textwidth]{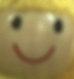} &
			\includegraphics[width=0.15\textwidth]{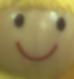} &
			\includegraphics[width=0.15\textwidth]{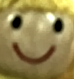} &
			\includegraphics[width=0.15\textwidth]{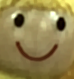} &	
			\includegraphics[width=0.15\textwidth]{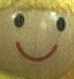} &
			\includegraphics[width=0.15\textwidth]{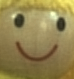} \\		
			
			\includegraphics[width=0.15\textwidth]{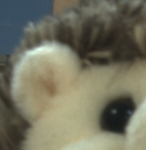} &
			\includegraphics[width=0.15\textwidth]{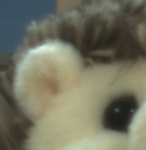} &	
			\includegraphics[width=0.15\textwidth]{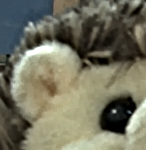} &
			\includegraphics[width=0.15\textwidth]{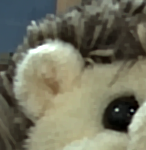} &
			\includegraphics[width=0.15\textwidth]{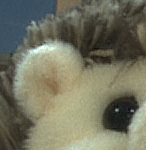} &
			\includegraphics[width=0.15\textwidth]{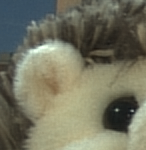} \\			
			
			\includegraphics[width=0.15\textwidth]{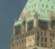} &
			\includegraphics[width=0.15\textwidth]{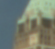} &
			\includegraphics[width=0.15\textwidth]{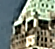} &
			\includegraphics[width=0.15\textwidth]{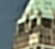} &
			\includegraphics[width=0.15\textwidth]{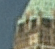} &
			\includegraphics[width=0.15\textwidth]{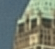} \\
			
			\includegraphics[width=0.15\textwidth]{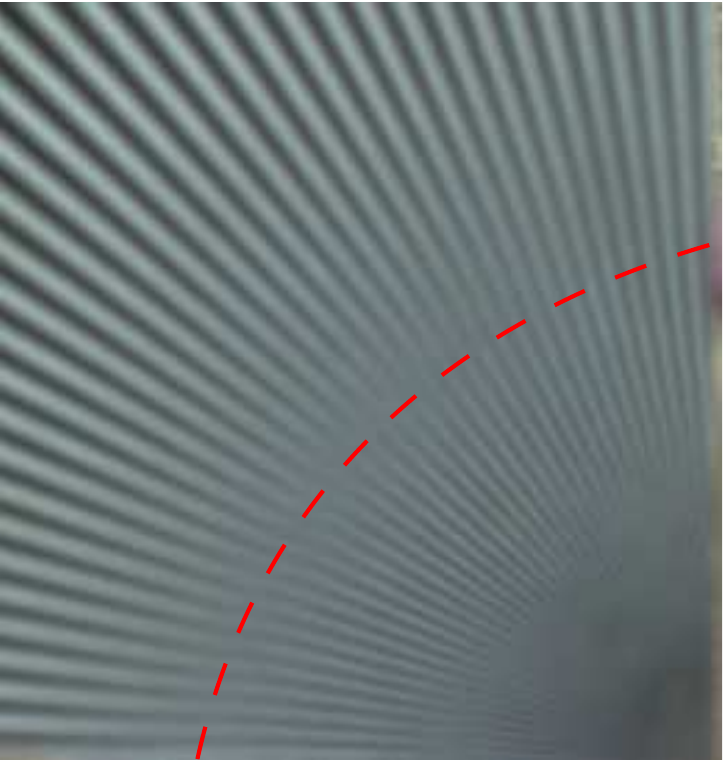} &
			\includegraphics[width=0.15\textwidth]{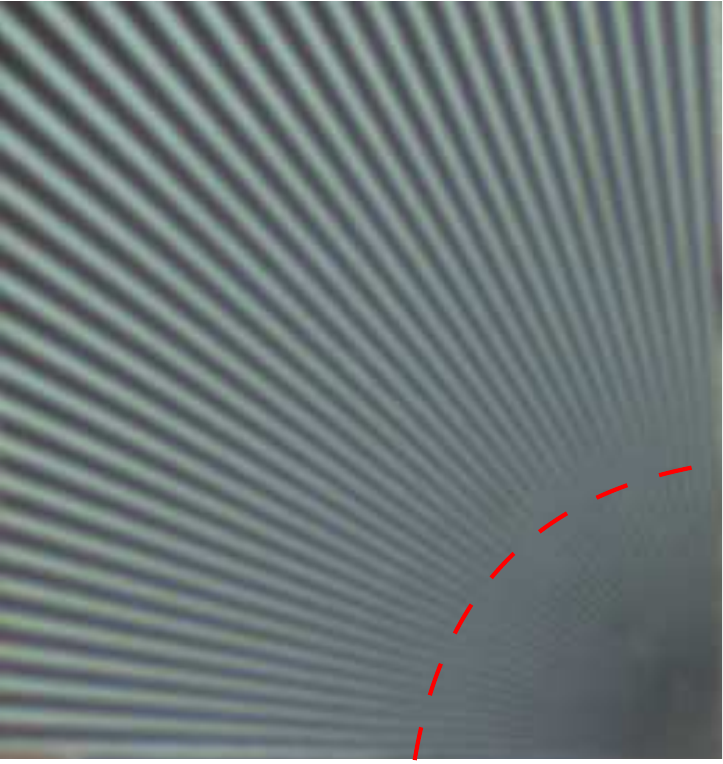} &		
			\includegraphics[width=0.15\textwidth]{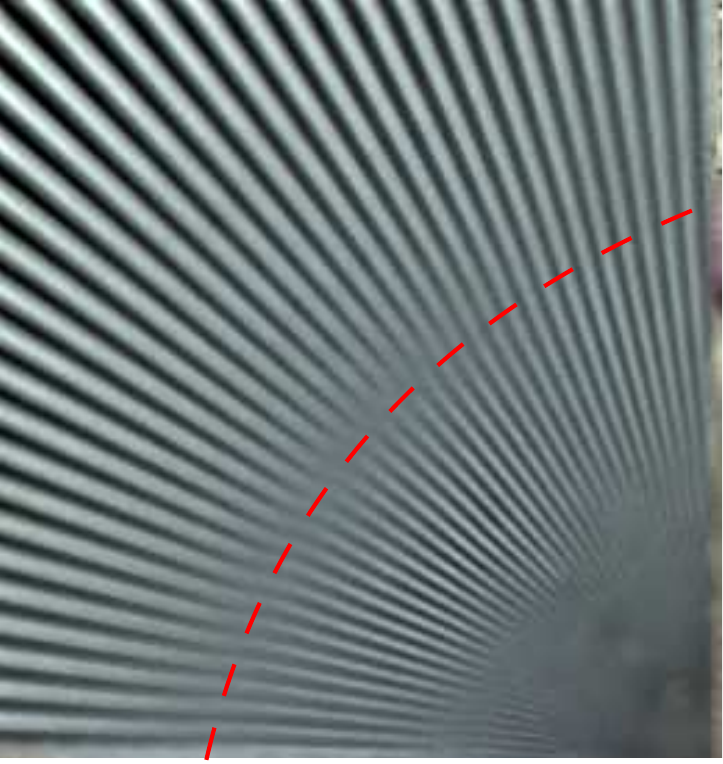} &		
			\includegraphics[width=0.15\textwidth]{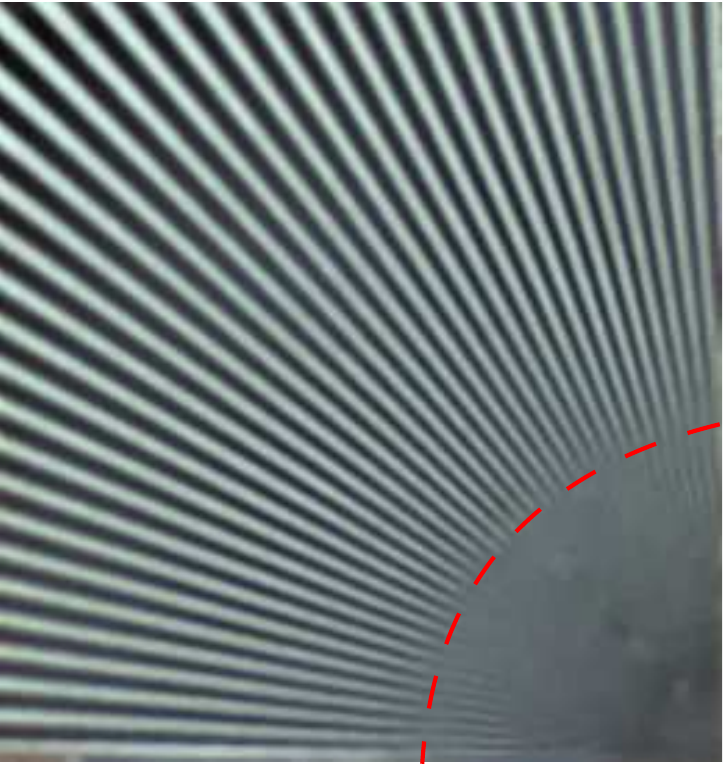} &		
			\includegraphics[width=0.15\textwidth]{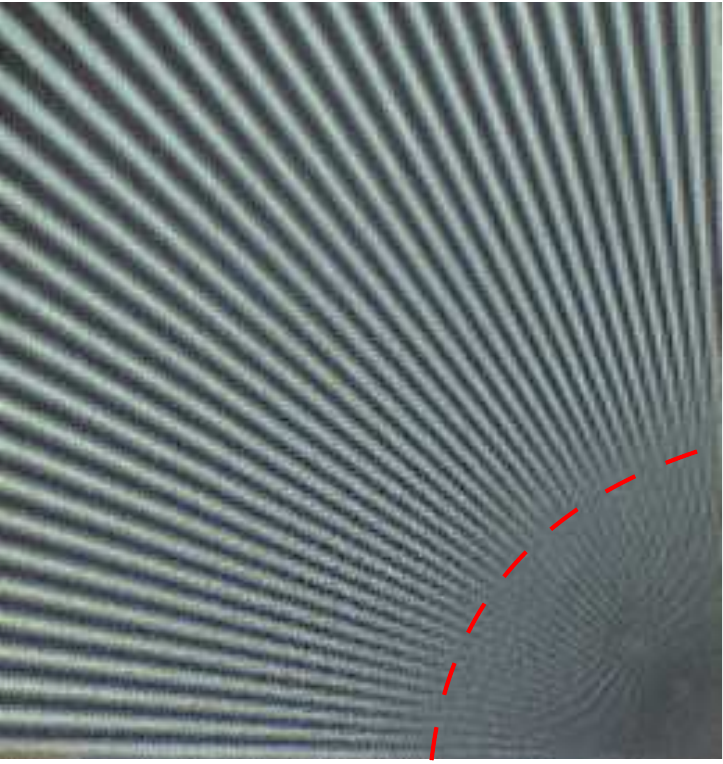} &
			\includegraphics[width=0.15\textwidth]{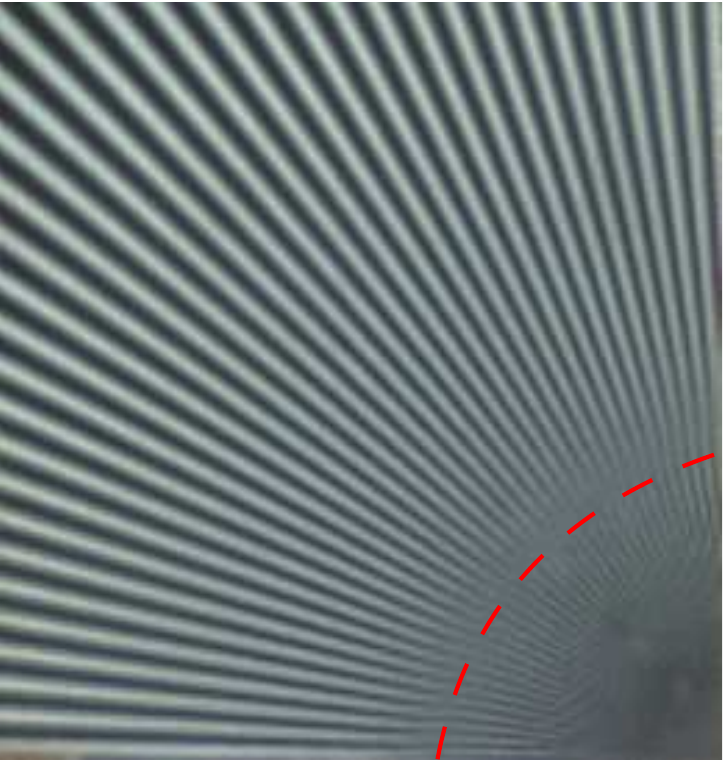} \\
			
			\multicolumn{1}{ c }{Clear} &
			\multicolumn{1}{ c }{Phase} &
			\multicolumn{1}{ c }{Clear + \cite{krishnan2011blind}} &		
			\multicolumn{1}{ c }{Phase + \cite{krishnan2011blind}} &		
			\multicolumn{1}{ c }{Phase + \cite{haim2014multi}} &			
			\multicolumn{1}{ c }{Phase + our } \\

			\multicolumn{1}{ c }{} &
			\multicolumn{1}{ c }{} &
			\multicolumn{1}{ c }{$450$ [sec]} &		
			\multicolumn{1}{ c }{$460$ [sec]} &		
			\multicolumn{1}{ c }{$140$ [sec]} &			
			\multicolumn{1}{ c }{$0.5$ [sec]} \\
			
		\end{tabular}   \\
		\caption{\red{\small \textbf{Full hardware pipeline experiment.} From left to right: clear aperture image, phase-coded aperture image, clear aperture image after deblurring using \cite{krishnan2011blind}, phase-coded aperture after deblurring using \cite{krishnan2011blind}, phase-coded aperture after deblurring using \cite{haim2014multi} and the output of our camera using a fixed-point neural network with $T=4$ layers. The dashed red lines in the bottom row indicate vanishing contrast. The reconstruction time required by \cite{krishnan2011blind} was $450$ seconds, by the OMP method from \cite{haim2014multi} $140$ seconds while our FPGA only requires $0.5$ seconds.}}
		\vspace*{-0.3cm}
		\label{table_DSLR_images}
	\end{figure}

	\red{A comparison between a conventional camera and our end-to-end system is presented in Figure \ref{table_DSLR_images}. The first and second columns from the left show the captured scene with a conventional lens (clear aperture) and with a phase-coded aperture respectively, with the focus set to the background poster. The third and fourth columns show the result of the blind deblurring algorithm from \cite{krishnan2011blind} applied to the clear and phase-coded aperture images, respectively. The fifth column presents the scene reconstructed from a phase-coded image using the OMP method from \cite{haim2014multi}. The rightmost column presents the reconstructed scene using our neural network with $4$ layers implemented on an FPGA. Both clear and coded-aperture images were captured in the exact same lighting conditions and exposure time. The entire image regions are shown in the top row whereas the four bottom rows contain zoomed-in snippets of the images at different depths. The superiority of the proposed system is evident in all cases in both quality and run-time. For example, in the magnified fragment of the background (to which the camera was focused), our system produces insignificant changes, while the deblurred clear and coded aperture images show significant over-sharpening artifacts visible as halos around the buildings and cartoon like segments. Our method also outperforms the OMP algorithm proposed by \cite{haim2014multi} which yields a noisier reconstruction. The bottom row presents a fragment of the resolution chart in which the vanishing contrast spatial frequency was marked with a dashed red line.}

	\section{Conclusions}
	\label{Conclusions}
	The phase-coded aperture computational EDOF imaging system presented in this work aims at solving one of the biggest challenges in today's miniature digital cameras, namely, acquisition of images with both high spatial resolution and large depth of field in demanding lighting conditions. Our proposed solution can be easily incorporated to
	existing imaging systems since it requires the addition of a thin mask (which can be integrated into one of the existing optical surfaces), and a simple real-time hardware computational unit, which was demonstrated in an FPGA prototype. We showed experimental evidence of the superior image quality of the proposed system compared to conventional acquisition and post-processing techniques. As we demonstrated through extensive experiments, our system outperforms existing techniques for post processing an image taken by a standard camera. It is worthwhile noting that the FPGA system developed for this project is rather flexible and, with moderate adjustments, may be used for implementing other image processing tasks based on sparse prior, such as denoising or restoration from compressed samples.  
		
	\bibliographystyle{splncs}
	\bibliography{egbib}
\end{document}